\documentclass[twoside]{article}
\usepackage{ecj,palatino,epsfig,latexsym,natbib,amsmath}
\usepackage{booktabs} 
\usepackage{standalone, tikz}
\usepackage{multirow}
\usepackage{ebproof}
\usepackage{amssymb} 
\usepackage{tikz-qtree}
\usepackage{syntax}
\usepackage{algpseudocode}
\usepackage{algorithm}
\usepackage{romannum}
\usepackage{soul}
\usepackage{array}
\usepackage{float}
\usepackage{placeins}
\usepackage{textgreek}
\usepackage{mleftright}

\newcommand*{\rom}[1]{\expandafter\@slowromancap\romannumeral #1@}
\newcolumntype{P}[1]{>{\centering\arraybackslash}p{#1}}

\begin{document}
\ecjHeader{30}{2}{291-327}{2022}{Modular GE for the Generation of ANNs}{K. Soltanian, A. Ebnenasir and M. Afsharchi}

\def\toptitlebar{ 
\noindent \textit{This version is the last pre-publication document accepted by ECJ, which has been published under volume 30, issue 2 on 2022.}
\hrule height.25pt
\vskip .25in} 

\title{\bf Modular Grammatical Evolution for the Generation of Artificial Neural Networks} 

\author{\name{\bf Khabat Soltanian} \hfill \addr{k.soltanian@znu.ac.ir}\\ 
\addr{Department of Electrical and Computer Engineering, University of Zanjan, 
Zanjan 45371-38791, Iran}
\AND
\name{\bf Ali Ebnenasir} \hfill \addr{aebnenas@mtu.edu}\\ 
\addr{Department of Computer Science, Michigan Technological University, 
Houghton MI 49931, USA}
\AND
\name{\bf Mohsen Afsharchi} \hfill \addr{afsharchi@znu.ac.ir}\\ 
\addr{Department of Electrical and Computer Engineering, University of Zanjan, 
Zanjan 45371-38791, Iran}
}

\maketitle

\begin{abstract}

This paper presents a novel method,  called Modular Grammatical Evolution (MGE),  towards validating the hypothesis that restricting the solution space of NeuroEvolution to modular and simple neural networks enables the efficient generation of smaller and more structured neural networks while providing acceptable (and in some cases superior) accuracy on large data sets.  MGE also enhances the state-of-the-art Grammatical Evolution (GE)  methods in two directions.  First, MGE's representation is modular in that each individual has a set of genes, and each gene is mapped to a neuron by  grammatical rules.  Second,  the proposed representation mitigates two important drawbacks of GE, namely the low scalability and weak locality of representation, towards generating modular and multi-layer networks with a high number of neurons. We define and evaluate five different forms of structures with and without modularity using MGE and find single-layer modules with no coupling more productive.  
Our experiments demonstrate that modularity helps in finding better neural networks faster.
We have validated the proposed method using ten well-known classification benchmarks with different sizes, feature counts, and output class counts. 
Our experimental results indicate that MGE provides superior accuracy with respect to existing NeuroEvolution methods and returns classifiers that are significantly simpler than other machine learning generated classifiers. 
Finally, we empirically demonstrate that MGE outperforms other GE methods in terms of locality and scalability properties.

\end{abstract}

\begin{keywords}
Grammatical Evolution;
Modular Representation;
NeuroEvolution.
\end{keywords}

\section{Introduction}

Despite the importance of Artificial Neural Networks (ANNs) in a variety of applications (e.g., pattern recognition, time series forecasting, and control systems) \citep{T-Kim, Miikkulainen17, Cantu, Xiao, Stanley2002}, the automated design of ANNs that have high learning capability and generalization, yet requiring low computational cost is a challenging problem. Designing ANNs involves two sub-problems, namely topology design and weight training. Topology design entails input feature selection, determining the number of hidden layers and hidden neurons along with connecting neurons. Weight training concerns with determining the weights of the links between neurons towards increasing learning and testing efficiency. Due to the practical significance of accurate and efficient ANNs, it is desirable to develop algorithmic methods that can synthesize smaller and more structured ANNs, especially for  resource-constrained devices.

Most existing methods for the design of ANNs are either semi-automated or lack sufficient machinery to simultaneously tackle all aspects of topology design and weight training. For example, \cite{Rumelhart} use back propagation errors to train the weights, but their method may get trapped in local optima and is applied usually to a manually designed neural network. \cite{kitano90} uses Evolutionary Algorithms (EAs) to solve the topology design problem. \cite{whitley89} also employs EAs to tackle the weight training problem. Various studies use EAs to simultaneously solve topology design and weight training in a fully automatic fashion \citep{Angeline93, Stanley2002}. 
\cite{Stanley2002} present NeuroEvolution of Augmenting Topologies (NEAT), a method based on the Genetic Algorithm (GA) where they use special data structures to codify the components of ANNs in genotype and apply user-defined crossover and mutation operators. Neural Architecture Search (NAS) uses different approaches like EAs, reinforcement learning, and Bayesian optimization to find near-optimal architectures for 
Deep Neural Networks (DNNs) \citep{Miikkulainen17, Zoph2018, Zhichao}. 
Genetic Programming (GP) and Grammatical Evolution (GE) \citep{Ryan1998} enable solution generation and optimization simultaneously \citep{Koza1994, Ryan1998, Miller2011}, which makes them suitable candidates for simultaneous generation and training of neural networks  \citep{tsoulos2008, ahmadizar2015, Assuncao2017, Assunco-gecco, Miller2019}. In particular, 
GE \citep{Ryan1998} is a variant of GP that takes a user-defined Context Free Grammar (CFG) as its input to determine how genotypes are mapped to phenotypes. The use of a CFG for encoding provides a high degree of user friendliness and flexibility for human designers in specifying  how EAs should function.   To enable the automated generation of lightweight and accurate ANNs for resource-constrained devices,  in this paper, we take the first step in validating the following overarching hypothesis. 




\begin{quote}
    {\it Restricting the solution space of NeuroEvolution to  simpler neural networks with a form of modularity 
increases the efficiency of these algorithms towards providing accuracy for large data sets, yet with smaller and more structured networks in terms of the number of neurons and topological structure.}
\end{quote}





This hypothesis is (i) inspired by previous studies \citep{Gasser, Chen2015, Miikkulainen17} demonstrating that structural modularity may increase the performance of neural networks while decreasing  network complexity, and (ii) motivated by the conjecture that modular construction of ANNs can limit the solution space for finding efficient networks faster.  To validate this hypothesis, we propose a  variant of GE, called Modular Grammatical Evolution (MGE).
MGE is modular at both the genotype and phenotype levels.
MGE's representation is modular in that it decomposes a solution to a set of simpler sub-solutions where a genotype encodes the sub-solutions as distinct genes. MGE also  considers a one-to-one correspondence between the genes and neurons as the most basic unit of computation in a network. In phenotype space, MGE  restricts the solution space to network topologies with a particular modular structure. Then, MGE  composes  neurons and modules towards generating  multi-layer and large ANNs with arbitrary topologies. 

\newpage
The {\bf contributions} of this paper are as follows:
\begin{enumerate}

\item A new representation that makes the encoding of a desired network structure more controllable in the evolutionary process. Such a representation enables (i) the  encoding of genes with a  one-to-one correspondence with neurons, and (ii) the modular composition of neurons at the encoding level. The proposed representation enables the propagation of neurons in the population during the evolution, and enables the reuse of neurons integrating in an ANN.


\item An algorithmic method for assembling neurons and generating feed-forward ANNs with multiple outputs and arbitrary architectures.

\item A novel notion of modularity as a collection of neurons with particular structural properties that limits the solution space towards more efficient neural networks.

\item A modularity-guided NeuroEvolution for the generation of large ANNs.
This contribution would not be possible without addressing the weak locality and low scalability problems in GE-based methods. That is, 
NeuroEvolution  should be able to manipulate a module  without changing other parts of the neural network; i.e.,   strong locality.
Such a method should not have severe bias towards small neural networks; i.e., should have high scalability.



\end{enumerate}


We have empirically evaluated MGE with respect to the  state-of-the-art methods on the following classification benchmarks: Flame, Wisconsin Diagnostic Breast Cancer (WDBC), Ionosphere, Sonar, Credit German, Heart Disease, Wine, Handwritten Digits, Letter, and MNIST. We note that the last three benchmarks are large data sets that require ANNs with multiple outputs. Our experiments validate that MGE outperforms NEAT and OVA-NEAT in terms of classification accuracy (which is the percentage of correctly predicted examples), yet the generated networks are significantly smaller than OVA-NEAT's networks, and between 14\% to 37\% smaller than NEAT's networks on Letter and MNIST. While OVA-NEAT's generated network contains 337 neurons and obtains 80.6\% of classification accuracy on MNIST, MGE's product containing 64 neurons achieves 89.3\% of accuracy. Convolutional Neural Network (CNN), Fully-connected Neural Network (FNN), and Support Vector Machine (SVM) provide more than 95\% of accuracy, however, their classifiers are more complex than MGE-generated classifiers up to 3 orders of magnitude. As such, the networks generated by MGE may be more fit for resource-constrained devices. The code of our implementation and the materials related to our experiments are available at {\it https://github.com/Khabat/Modular-NeuroEvolution.git}.


\section{Basic Concepts: Grammatical Evolution}
\label{sec:basics}
\begin{figure}
\center
\includegraphics[width=0.70\textwidth]{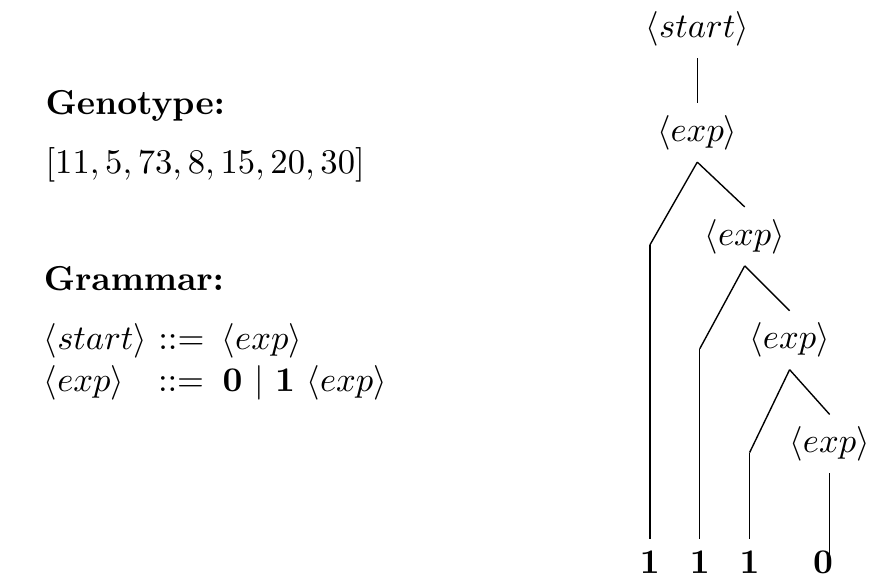}
\caption{Given a genotype and a CFG grammar (left), the derivation tree (right) shows how the genotype is mapped to its corresponding phenotype, ``1110'', using the given grammar. }
\label{fig:derivationtree}
\end{figure}

Grammatical Evolution (GE) evolves sentences in a language that is described by a Context-Free Grammar (CFG) in Backus-Naur Form (BNF). 
A BNF grammar is a tuple $\langle N, T, P, S \rangle$ where $N$, $T$, $P$, and $S$ respectively represent non-terminal or intermediate symbols, the set of terminal symbols, the set of production rules and the start symbol.
The grammar is used to map a genotype to a candidate solution in phenotype space, which is a sentence in the language of the CFG.
The genotype binary string is decoded as a sequence of \textit{8-bit} integer codons that are read from left to right in the genotype-phenotype mapping process as follows:
A derivation tree is initiated with the start symbol as the root of the tree (See Figure 1). 
The derivation is performed by expanding non-terminal symbols until there is no non-terminal symbol in the leaves of the tree.
If the production rule corresponding to a non-terminal has multiple options, the next codon of genotype will be read and used to choose which one to fire.
The codon value modulo the number of expansion possibilities for the current production rule determines the option that fires. 
In each step, if there are multiple non-terminals in the derivation tree, the left-most of them is expanded. 
If the process continues while the end of the genotype is reached, then the codons are read from the start of the genotype again, called \textit{wrapping}. 
If after a predefined number of wrapping, the mapping process does not terminate successfully, the mapping is halted while the individual will get the lowest fitness score. We label these individuals as {\it invalid} ones.
For example, a simple BNF grammar denoted $G_1$ with two production rules is as follows:
\begin{grammar}
<start> ::= <exp> 

<exp> ::= \textbf{0} | \textbf{1} <exp> 
\end{grammar}
where, $N$=\{$\langle\hspace{0pt}start\hspace{0pt}\rangle$, $\langle\hspace{0pt}exp\hspace{0pt}\rangle$\}, $T$=\{0,1\}, and $S$=\{$\langle\hspace{0pt}start\hspace{0pt}\rangle$\}.
$G_1$ generates sentences starting with any number of consecutive \textit{1}s and terminating with \textit{0}.
Figure \ref{fig:derivationtree} demonstrates a simple genotype and the derivation tree generated by $G_1$ in genotype-phenotype mapping. 
The output of the mapping is ``1110", while the input, known as genotype, is assumed to be: [11, 5, 73, 8, 15, 20, 30].
The start symbol $\langle\hspace{0pt}start\hspace{0pt}\rangle$ expands to $\langle\hspace{0pt}exp\hspace{0pt}\rangle$ with no codon use because it has only one possible production. 
Then first codon, 11, is used to expand the first $\langle\hspace{0pt}exp\hspace{0pt}\rangle$ symbol as follows: 11 modulo 2 equals to 1 which results in selecting the second option for expansion (2 = \textit{the number of options for $\langle\hspace{0pt}exp\hspace{0pt}\rangle$}). 
The process continues until reaching a valid sentence that contains only terminal symbols. 
Notice that, a part of the genotype (15, 20, 30) was not used for the mapping. 

\section{Problem Statement}
\label{sec:problemstatement}

In order to validate the underlying hypothesis of this work (stated in the Introduction), we pose the following Research Questions (RQs):






\begin{enumerate}
\item {\bf RQ1}: Since the most basic unit of abstraction/computation in an ANN includes a neuron, it would be useful to have fine-grain control over how the neurons are connected with each other throughout the evolution. Thus, we would like to know: {\em Are there  representations that enable (i) the  encoding of genes with a  one-to-one correspondence with neurons, and (ii) the modular composition of neurons at the encoding level}? Such representations will make the encoding of a desired network structure more controllable in the evolutionary process whereas most existing methods are monolithic; i.e., they encode the entire ANN as a whole.

\item {\bf RQ2}:  {\em  Is there a method for enabling  the encoding of feed-forward ANNs with multiple outputs}?

\item {\bf RQ3}: {\em How can we assemble neurons during the evolutionary process  towards creating arbitrary network architectures}?

\item {\bf RQ4}: {\em What is the meaning of a module in the phenotype space}? and {\em What kind of structure should a module have so that the composition of modules results in an ANN that provides better accuracy, preferably with a smaller and more modular structures}?

\end{enumerate}

GE's literature contains many efforts and debates to tackle some of the problems in its genotype-phenotype mapping \citep{Rothlauf, Oniel99, Galvan-Lopez, Lourenco, Medvet2017, Medvet17, Assunco-gecco, Bartoli, Medvet2019}. To address the aforementioned questions, we also improve the GE method in two fronts: locality and scalability. We address the limitations of existing methods regarding these two properties, thereby enabling the generation of scalable and modular ANNs.

\section{Related Work}
\label{sec:relatedwork}
In this section, we discuss existing methods and classify them based on the RQs stated in Section \ref{sec:problemstatement}.

\subsection{Representation Design}
The invalidity, as the rate of producing invalid genotypes by random initialization or genetic operators is inevitable in the original GE. 
Locality is the other important aspect of any representation that relates the neighborhood structure of genotype space to that of phenotype space. 
In a strong locality, genotypic neighbors would be neighbor in phenotypic space, too. 
The GE representation suffers from the weak locality problem because a bit flip (in genotype) that corresponds to a derivation step near the root of the derivation path could result in a completely different phenotype \citep{Rothlauf, Lourenco}. 
For example, in Figure \ref{fig:derivationtree}, if we change the first codon of the genotype to `\textbf{4}', (to have the genotype: [\textbf{4}, 5, 73, 8, 15, 20, 30]) the generated string would be `0' instead of `1110'. 
Experimental studies on evolutionary algorithms (including GE-variants) show that stronger locality makes better performance \citep{Galvan-Lopez, Lourenco, Medvet2017, Medvet17, Gottlieb, Nguyen, Rothlauf}. 

Researchers propose various representations (or mapping approaches) as variants of GE to overcome the aforementioned issues with GE's mapping.
For example, a genotype in Structured Grammatical Evolution (SGE) \citep{Lourenco} includes $|N|$ lists or sub-strings that are called genes where \textit{N} is the set of non-terminals in grammar \textit{G}.
During the mapping process, if a non-terminal $p \in N$ appears in derivation tree, the leftmost unused codon in its corresponding gene is used to expand $p$, 
while in GE the leftmost unused codon in the genotype is used. 
The partitioning partly addresses the weak locality problem \citep{Lourenco};
still, when the gene of a non-terminal \textit{p} changes, the derivation tree of successor non-terminals of \textit{p} change, too.
Moreover, SGE evolves fixed size genotypes whereas it is variable in GE that makes GE more flexible. 
SGE's representation solves the invalidity problem, at the expense of increasing the length of the genotype.
Dynamic Structured Grammatical Evolution (DSGE) \citep{Assunco-gecco} uses a variable length genotype;
Non-terminals' genes grow as required, while in SGE genes length are fixed and calculated before evolution. 
DSGE avoids the invalid individuals via an extra repair mechanism added to the mapping procedure.
Regarding locality issue, SGE and DSGE behave similarly.
Weighted Hierarchical Grammatical Evolution (WHGE) \citep{Bartoli} uses a hierarchical tree like structure for partitioning the genotype and assigns the resulting sub-strings of genotypes to non-terminal symbols in the derivation tree. 
The method is called weighted because it uses the expressive power of non-terminals to assign them sub-strings of genotype. 
Recently, \cite{Medvet2019} propose a method to automatically design new genotype-phenotype mapping rules for GE, which exhibits stronger locality along with lower redundancy. 
Their experiments demonstrate that the search effectiveness of the generated representations compare successfully with human-designed representations (GE, improved GE, Hierarchical GE and WHGE).

MGE's representation and mapping are different from all of the aforementioned methods in several directions. 
First, the modules of a solution are encoded separately in the genotype while other methods encode all of the solution's components at once (monolithically).
The direct outcome is that locality rises because when a module is perturbed other modules remain unchanged. 
Second, the proposed representation alleviates the invalidity problem without any extra process (e.g., the DSGE's repair mechanism) or limitation (e.g., fixed length genotype in SGE and WHGE). 
MGE decodes the modules independently, that is if some modules' derivation fails other modules compose into a feasible solution.
Third, the new representation allows defining tailored crossover and mutation operators that make the evolutionary engine more controllable than other GE variants. 

\subsection{Scalability}
\label{sec:scalability}
The other issue with the mapping of GE-based methods is the low scalability. 
The GE's mapping process has an intense bias towards smaller solutions than bigger ones that makes it nearly impossible for very big structures to be generated \citep{Thorhauer}.
For example, in the grammar $G_1$ of Figure \ref{fig:derivationtree}, there must be $k$ consecutive codons with odd integers that are followed by an even codon if we want to have a sentence with $k$ consecutive `1's that terminates with a `0' symbol. 
As the codons are filled with random integers in the range [0, 255], the probability for such an artifact is $\frac{1}{2^{k+1}}$. 
If we assume the genotype contains 7 codons without wrapping mechanism, the possible sentences would be: ``0", ``10", ``110", ``1110", ``11110", ``111110", ``1111110", or returns ``invalid individual" while their probabilities are respectively 0.5, 0.25, 0.125, 0.0625, 0.03125, 0.0125625, 0.0078125, and 0.0078125. 
Observe that, the probability decreases exponentially when the size of the artifact grows linearly. 
This issue in GE, makes it less probable to find complex solutions as demonstrated in Section \ref{sec:repanalysis}.
MGE addresses the problem where the modular genotype permits to encode any arbitrary number of modules in the solution.

\subsection{Modularity}
\label{sec:MinEA}
A modular solution contains almost independent units (modules) where each unit plays a specific role in solution's performance \citep{Amer2019}. 
Modularity provides important advantages: reuse, readability, and scalability \citep{Koza1994, Swafford2011}. 
While the modules may be highly complex internally, they are loosely connected together usually in a modular solution. 
\cite{angeline1993, angeline1992} present the first method adding modularity to  evolutionary algorithms where the augmented reproduction process can \textit{group} a portion of an offspring as a module to protect it from future manipulations or \textit{ungroup} a module to make it part of the solution again.  

There are several studies that investigate the importance of modularity in the performance of GPs \citep{Koza1994, Miller2011, Ellefsen2019, Oneill2000, Hemberg2009, Harper, Swafford2011, Swafford2011a}. 
For example, \cite{Koza1994} demonstrates that adding modularity to GP highly increases scalability. 
Cartesian Genetic Programming (CGP) \citep{Miller2011} is one of the well-known modular GPs that applies to many applications like circuit design, GPU programming, ANN synthesis \citep{Miller2019}, and DNNs design \citep{Ellefsen2019}. 
CGP is modular in that it represents sub-functions of the target function as distinct sub-trees in predefined layers where the output sub-tree composes the final candidate solution. 
However, MGE and CGP are different in two major ways: (I) MGE is based on grammatical representation not tree encoding, 
(II) MGE automatically generates  the structure of the solution (the number and position of modules) while the user should pre-define it for CGP.

Some existing experimental results demonstrate significant improvement in the performance of Grammatical Evolution, if a form of modularity is added \citep{Harper, Hemberg2009}.
Other examples of incorporating modularity in GEs include \citep{Oneill2000, Harper, Swafford2011, Swafford2011a}. 
Adding functions to the grammar is among the earliest studies that investigate  modularity in GE \citep{Oneill2000}. 
\cite{Harper} propose a different approach where the grammar is evolved along with the solution simultaneously. 
However, the modules are not independent that is a change into a module's region in the genotype can alter the entire phenotype. 
Likewise, the modules are unable to propagate in the population. 
\cite{Swafford2011a} propose another approach that augments the grammar during the evolutionary process with useful modules.
However, \cite{Swafford2011a} must use the heuristic rules for determining good modules, and require extra computational cost for adding the discovered modules to the grammar.
MGE provides  modularity in genotype level where the genetic operators are able to refine the modules, generate new modules, and propagate the good modules across the population.
Moreover, MGE enables a module's output to be reused by other modules. 
There are also studies that examine the role of modularity in ANN's performance for some specific problem domains, such as complex problems with multiple tasks \citep{Chandra2018}, problems that contain various independent sub-problems \citep{Khare2005, Ellefsen2015}, and problems with heterogeneous inputs \citep{Schrum}. 
Instead, our algorithm imposes a form of modularity to the target neural networks for classification problems.


\subsection{NeuroEvolution}
\label{sec:GENE}
\cite{Anil2018} train the weights of single-layered neural networks using cooperative co-evolutionary differential evolution for classification benchmarks.
NEAT evolves both the topology and weights of a neural network and returns significant results on different pole-balancing benchmarks. 
Due to efficiency and well-defined crossover operator of NEAT, several researchers adopt NEAT for other types of machine learning tasks; 
\cite{Grisci} use a variant of NEAT \citep{Whiteson} for feature selection in micro-array data with over 50,000 features, 
\cite{McDonnell} develop ensemble methods based on NEAT to solve fairly complex classification problems, 
and \cite{Alexander} extend NEAT to evolve activation functions besides network topology and connection weights for three classic classification and regression problems. 
In Blocky-Net \citep{Watts}, genetic algorithm evolves the entire neural network for classification problems. 
\cite{Ellefsen2019} guide NeuroEvolution by employing structure-related objectives plus the main objective for two decomposable tasks: retina problem, and robot locomotion.

To the best of our knowledge, \cite{tsoulos2008} propose the first NeuroEvolution method using GE where they generate the topology and weights of feed-forward neural networks with one hidden layer. 
\cite{ahmadizar2015} propose a new representation that encodes the weights of neural networks in real number format, 
while it encodes the topology of the network with GE's representation. 
\cite{Assuncao2017} use SGE to evolve single-layered neural networks. In another work, Assun\c{c}\~{a}o et al. utilize DSGE to evolve multi-layer neural networks and obtain better results than the results that  SGE returns \citep{Assunco-gecco}.
There are also other grammar-based methods than GE that generate neural network topology through developmental rules \citep{kitano90, Gruau}.
Two major novelties of MGE with respect to other NeuroEvolutions in addition to the algorithmic and encoding schema listed in this section is that, (i) MGE takes the advantage of a structural constraint (modularity) on the networks to obtain simpler neural networks with higher generalization ability more efficiently and (ii) we approach the relatively larger classification tasks with multiple classes.

\section{MGE}
\label{sec:MGE}
This section addresses the research questions RQ1, RQ2 and RQ3 stated in Section \ref{sec:problemstatement}. The main steps of the evolutionary algorithm that MGE performs are as following:
\begin{enumerate}
\item \textit{Initialization} (generate random candidate solutions)
\item \textit{Evaluation} (map the genotypes to phenotypes, then calculate their fitness using the problem's dataset)
\item \textit{Parent Selection} (randomly select and move individuals to the mating pool considering their fitness) \label{case:three}
\item \textit{Crossover Operator} (the offspring population emerges of recombination of the parents pairs)
\item \textit{Mutation Operator} (slightly change the offspring genotype by random perturbations)
\item \textit{Evaluation} (map the genotypes to phenotypes, then calculate their fitness using the problem's dataset)
\item \textit{Survival Selection} (select individuals among current population and offspring population to appear in the next generation)
\item Go to Step \ref{case:three} for a pre-defined generation count
\item Return the \textbf{best individual} of last generation
\end{enumerate}
The main distinctive feature of our algorithm includes the representation we define. Moreover, the mapping and genetic operators of MGE are different from other algorithms like GE. Next, we explain the proposed representation, mapping process and genetic operators with examples. 

\subsection{Representation}
\label{sec:rep}
The proposed representation (in response to RQ1) is able to encode ANNs with almost any arbitrary architecture. In this paper, the networks are restricted to feed-forward ANNs with multiple output neurons (where networks with single output neuron are generated for binary class datasets). 

\noindent{\bf Gene to Neuron Mapping}.\ 
According to our method, for each hidden neuron in the ANN there is a corresponding gene in genotype. 
Hence, a genotype is only a list of neuron genes.
In fact, a gene encodes the connectivity and weights of a neuron in standard grammatical representation, while previous methods (GE, SGE, and DSGE) encode the entire ANN in one package. 

\begin{figure*}[h]
\includegraphics[width=1\textwidth]{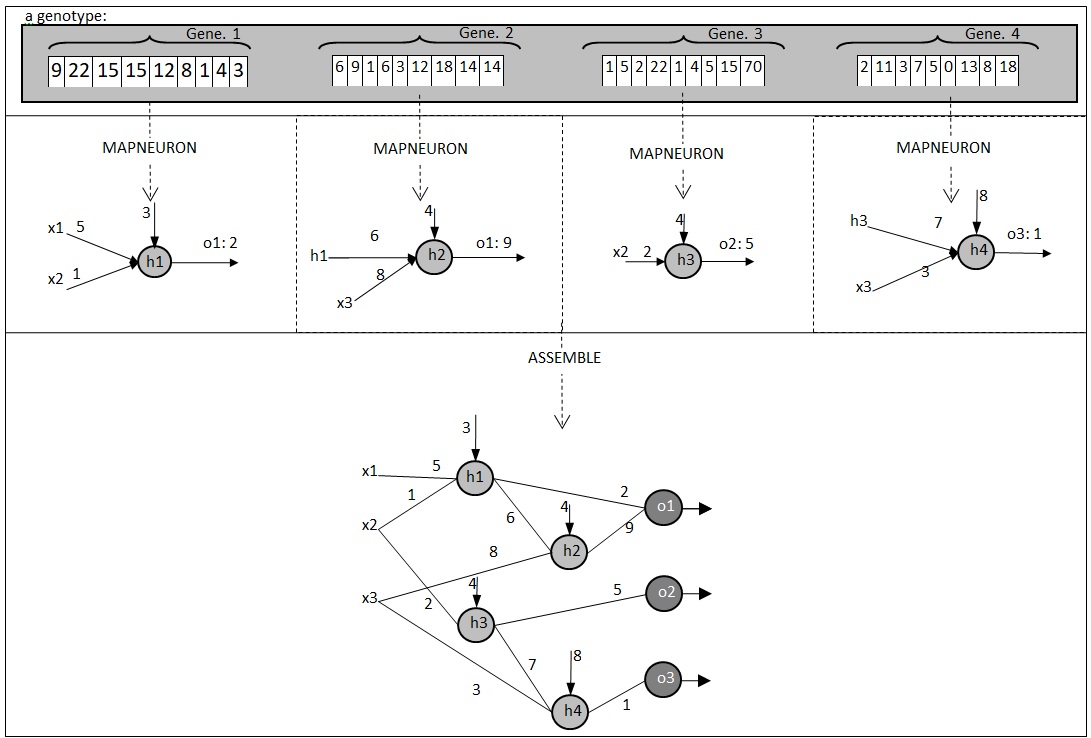}
\caption{A simple genotype and the main mapping steps to generate the corresponding phenotype (a multi-layer neural network with four hidden neurons and three output neurons). The genotype contains four genes, where the first gene is mapped to a neuron, \textit{h1} via GE mapping, then after adding the reference of the mapped neuron, \textit{h1}, to the mapping grammar the second gene is mapped, and the process continues for third and fourth genes. The last step, shows a composition of these neurons in a neural network with three output neurons.}
\label{fig:individual}
\end{figure*}

\noindent{\bf Assembling Neurons}.\ 
To make almost any arbitrary topologies possible, we assume that the hidden neurons (genes) are in distinct layers.
Such a structure allows a different mapping process where a neuron is able to feed from its prior mapped neurons.
However, the connections that a neuron makes to its prior neurons determine the actual layer of that neuron. 
For example, if a neuron makes no connection with the prior neurons, it will be regarded as in the first hidden layer of the neural network. 
Figure \ref{fig:individual} presents an example genotype with 4 genes and its corresponding neural network with 4 hidden neurons and 3 output neurons. 
In this example, input features include $x_1$, $x_2$, and $x_3$. 
It should be stated that, a gene encodes also the connections between the corresponding hidden neuron and the output neurons. 
Thus, the output neuron is only an accumulator of other neurons' outputs, which needs no gene. In subsequent sections, we shall explain the details of the developed BNF grammar and the process of mapping.

\subsection{Proposed Neuron-Generating Grammar}
\begin{figure}
\center
\includegraphics[width=0.8\textwidth]{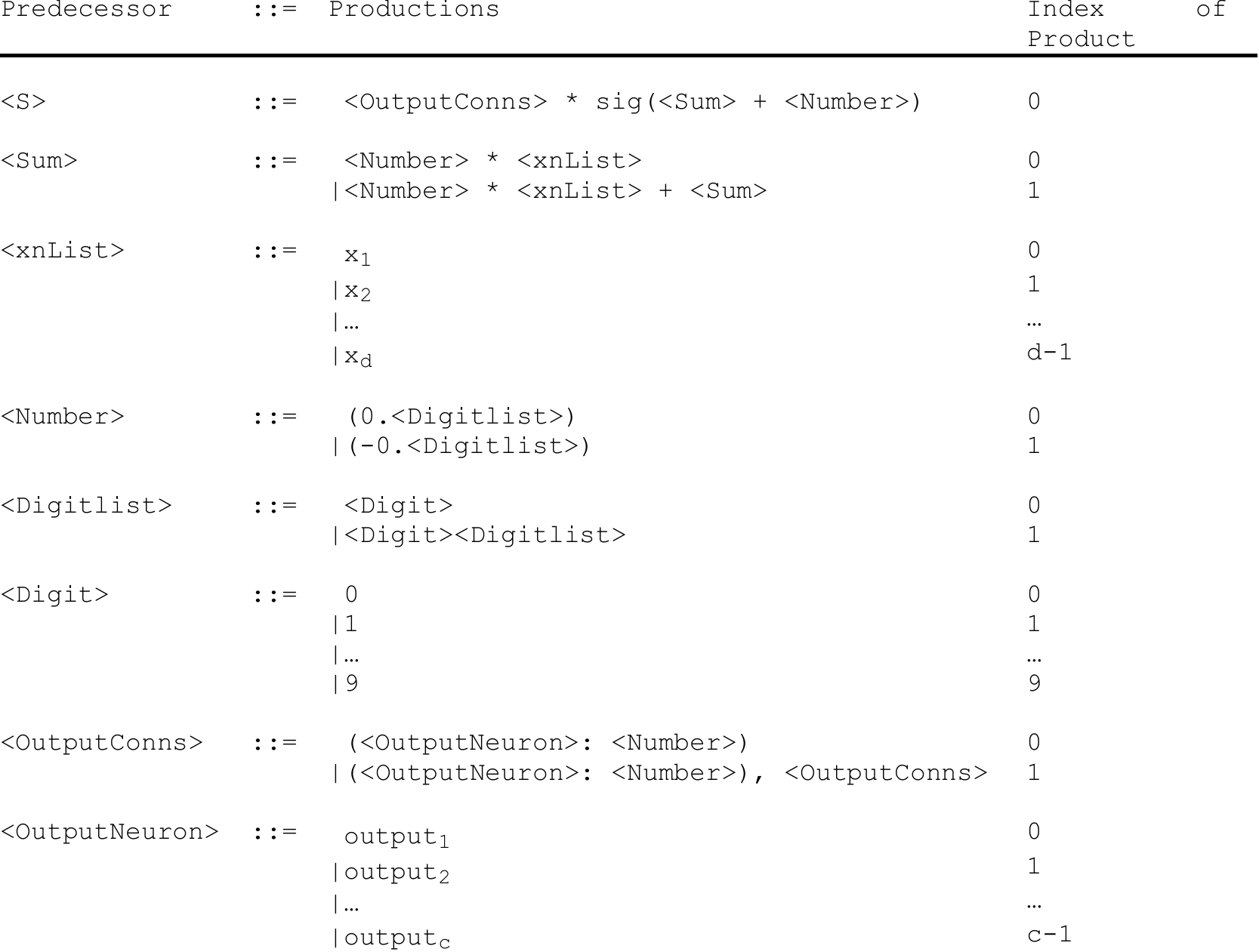}
\caption{The BNF grammar used to generate neurons.}
\label{fig:cfg2}
\end{figure}

\sloppy
The proposed grammar maps a gene, a vector of integer codons to a neuron with any arbitrary input weighted connection and output weighted connection to one or multiple output neurons of the network.
Figure \ref{fig:cfg2} demonstrates the production rules of the proposed CFG, described by the tuple $G_2=\langle\hspace{0pt}S, N, T, P\rangle\hspace{0pt}$.
The grammar generates a neuron including one or more input connections, a bias connection (that plays the role of neuron's activation threshold), and some direct connections to the output neurons of the network as well as the weights of the connections. 
We denote the number of input features and classes respectively with \textit{d} and \textit{c}.
Non-terminal $\langle\hspace{0pt}S\rangle\hspace{0pt}$ is the start symbol that represents a complete hidden neuron scheme. 
The set of non-terminals is $N = \{\langle\hspace{0pt}S\rangle\hspace{0pt}, \langle\hspace{0pt}Sum\rangle\hspace{0pt}, \langle\hspace{0pt}xnList\rangle\hspace{0pt}, \langle\hspace{0pt}Number\rangle\hspace{0pt}, \langle\hspace{0pt}Digitlist\rangle\hspace{0pt}, \langle\hspace{0pt}Digit\rangle\hspace{0pt}, \langle\hspace{0pt}OutputNeuron\rangle\hspace{0pt}, \langle\hspace{0pt}OutputConns\rangle\hspace{0pt}\}$ where the non-terminal $\langle\hspace{0pt}Number\rangle\hspace{0pt}$ represents a real number in the range of (-1, +1) to act as a connection weight, the non-terminal $\langle\hspace{0pt}xnList\rangle\hspace{0pt}$ captures an input connection, $\langle\hspace{0pt}OutputConns\rangle\hspace{0pt}$ generates weighted connection(s) with output neuron(s), and $\langle\hspace{0pt}OutputNeuron\rangle\hspace{0pt}$ is the index of the output neuron to which the neuron connects.
The \textit{sig} refers to a sigmoid activation function for hidden neurons calculated as $sig(x) = \frac{1}{1+e^{-x}}$. 
Initially, $\langle\hspace{0pt}xnList\rangle\hspace{0pt}$ simply contains dataset input features (terminals that are expressed as $x_1$, $x_2$, \dots, $x_d$).
However, it may change during mapping a genotype; when a gene is mapped, its neuron reference is added to the list temporarily. 
It allows next hidden neurons to get input connection from the mapped neurons. 
When the mapping of a genotype finishes, the rule resets to the initial state. 

\fussy 
\subsection{Mapping Genotype to Phenotype}
The mapping process transforms a genotype containing {\it h} genes into a neural network with up to {\it h} hidden neurons.
Algorithm 1 demonstrates how the genes are mapped to their corresponding neurons.
Line 2 initializes {\it neurons} as an empty list which is considered to hold the neurons generated via the mapping process.
Line 3 sets {\it neuronsCount} to zero as the counter of neurons that \textbf{for all} loop generates.
Line 4 sets {\it grammar'} to {\it grammar}, as {\it grammar'} will be updated during the mapping process while {\it grammar} should be preserved for next mappings.
In Lines 5-13, the \textbf{for all} loop maps the genes in the genotype as follows.
In Line 6, the \Call{MapNeuron}{} procedure, explained in Section \ref{sssec:mappinggene}, gets a gene and the grammar, then returns the gene's corresponding neuron ({\it neuron}).
However, if a gene's end is reached and still there is some non-terminals in the generated string, \Call{MapNeuron}{} returns a constant string, ``INVALID'' (we use no wrapping mechanism); in that case the gene is ignored by the algorithm in Line 7. 
Otherwise, Lines 8-11 execute.
Line 8 adds {\it neuron} to the {\it neurons} list, and Line 9 increments {\it neuronsCount}. 
Line 10 creates a reference to the generated neuron (e.g., {\it neuronIndex = ``h1"} for the first neuron), and Line 11 updates the {\it grammar'} by adding the reference (e.g., {\it ``h1"}) to the productions list of non-terminal $\langle\hspace{0pt}xnList\rangle\hspace{0pt}$.
If none of the genes is mapped successfully (Line 14), the algorithm returns ``invalid individual'' in Line 15.
Otherwise, if there is at least one valid neuron, the algorithm calls the  \Call{Assemble}{} procedure to create a neural network ({\it neuralNetwork}) by considering {\it c} output sigmoidal neurons that sum the outputs of the synthesized neurons (held in {\it neurons}).
Finally, the algorithm returns {\it neuralNetwork} (Line 18). 

As an example of mapping we can refer back to Figure \ref{fig:individual}.
Figure \ref{fig:individual} depicts that Gene1, Gene2, Gene3, and Gene4 are mapped to their corresponding neurons (\textit{h1}, \textit{h2}, \textit{h3}, and \textit{h4}) in order, and then the generated neurons are assembled in a neural network.
This order of mapping indicates that \textit{h2} can get connection from \textit{h1}, and {\it h3} can get connection from {\it h1} and {\it h2} and so on.

\begin{algorithm}
\label{alg:alg1}
\caption{Genotype to phenotype mapping procedure}
\begin{algorithmic}[1]
\Procedure {MapIndividual}{$genotype$, $grammar$}
\State $neurons \leftarrow []$
\State $neuronsCount \leftarrow 0$
\State $grammar' \leftarrow grammar$
\ForAll {$gene \in genotype$}
\State $neuron \leftarrow \Call{MapNeuron}{gene, grammar'}$
\If {$neuron \neq$ ``INVALID''}
\State $neurons \leftarrow neurons \cup neuron$
\State $neuronsCount \leftarrow neuronsCount+1$
\item[] \hspace*{38pt} \(\triangleright\)update grammar: (e.g., $\langle\hspace{0pt}xnList\rangle\hspace{0pt}:: x_1|...|x_d|\textbf{h1}$)
\State $neuronIndex \leftarrow \Call{ConCat}{``h",neuronsCount}$
\State $grammar'.\langle\hspace{0pt}xnList\rangle\hspace{0pt}.productions.\Call{Add}{neuronsIndex}$
\EndIf
\EndFor
\If {$neuronsCount = 0$}
\State \textbf{return} ``invalid individual''
\EndIf
\item[]
\item[] \hspace*{13pt}\(\triangleright\)put all the neurons together via some sigmoidal output neuron(s)
\State $neuralNetwork \leftarrow$ \Call{Assemble}{$neurons$} 
\State \textbf{return} $neuralNetwork$
\EndProcedure
\end{algorithmic}
\end{algorithm} 

\subsubsection{Mapping Genes to Neurons (MapNeuron)}
\label{sssec:mappinggene}
The inputs of the \Call{MapNeuron}{} includes a gene, and a neuron-generating CFG.
A gene, as a part of a genotype, is an array of integer values in range [0, 255] (e.g., Gene1 and Gene2 in genotype of Figure \ref{fig:individual}).
The output of the procedure is a neuron including its weighted input connections and weighted connections to the output neurons.
The output connection ensures that all the neurons affect the functioning of the network. 
\sloppy
The mapping process executes as follows.
The $\langle\hspace{0pt}S\rangle\hspace{0pt}$ symbol starts the process, while it is replaced with its sole product: $\{\hspace{0pt}\langle\hspace{0pt}OutputConns\rangle\hspace{0pt}$ * sig($\langle\hspace{0pt}Sum\rangle\hspace{0pt}$ + $\langle\hspace{0pt}Number\rangle\hspace{0pt}$)$\}\hspace{0pt}$.
Then, the left-most non-terminal in the string, $\langle\hspace{0pt}OutputConns\rangle\hspace{0pt}$, is replaced with one of its two products, $(\langle\hspace{0pt}OutputNeuron\rangle\hspace{0pt}:\hspace{0pt}\langle\hspace{0pt}Number\rangle\hspace{0pt})$ or $\langle\hspace{0pt}OutputConns\rangle\hspace{0pt}(\langle\hspace{0pt}OutputNeuron\rangle\hspace{0pt}:\langle\hspace{0pt}Number\rangle\hspace{0pt})$, based on the first codon of the gene. 
Since there are two products, the codon's value modulo 2 determines which one to substitute $\langle\hspace{0pt}OutputConns\rangle\hspace{0pt}$.
The process continues for remaining non-terminals in the expression using next codons of the gene until non-terminals disappear.
For ease of  presentation,  we describe all steps of a gene decoding as follows.

\begin{figure}
\center
\includegraphics[width=0.85\textwidth]{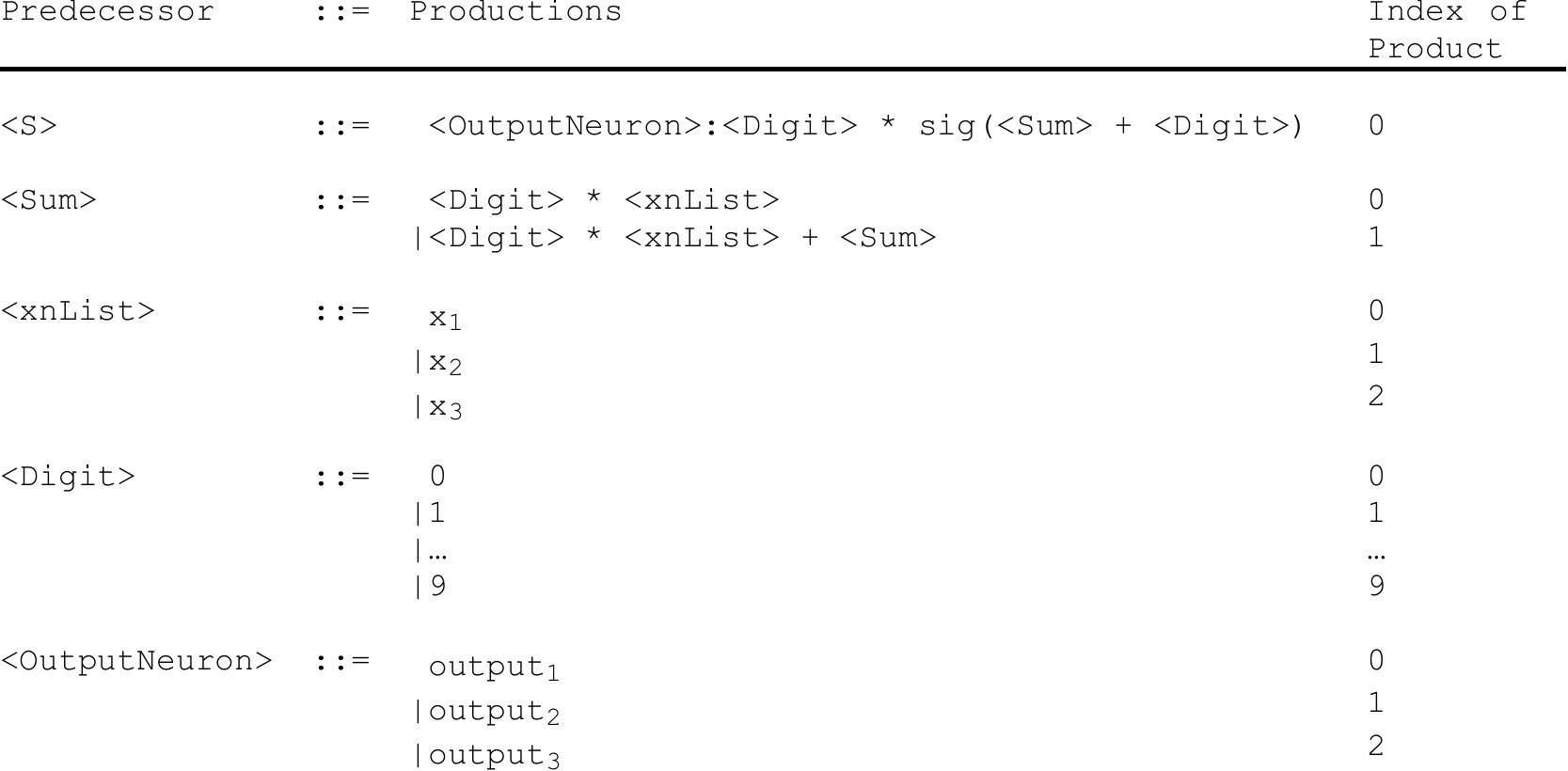}
\caption{A simplified version of the grammar proposed in Figure \ref{fig:cfg2}. It assumes that the produced neural networks have three output neurons, where each hidden neuron connects directly to exactly one output neuron and also the grammar assumes that connection weights are single-digit integers.}
\label{fig:cfg3}
\end{figure}

\begin{figure}[H]
\center
\includegraphics[width=0.78\textwidth]{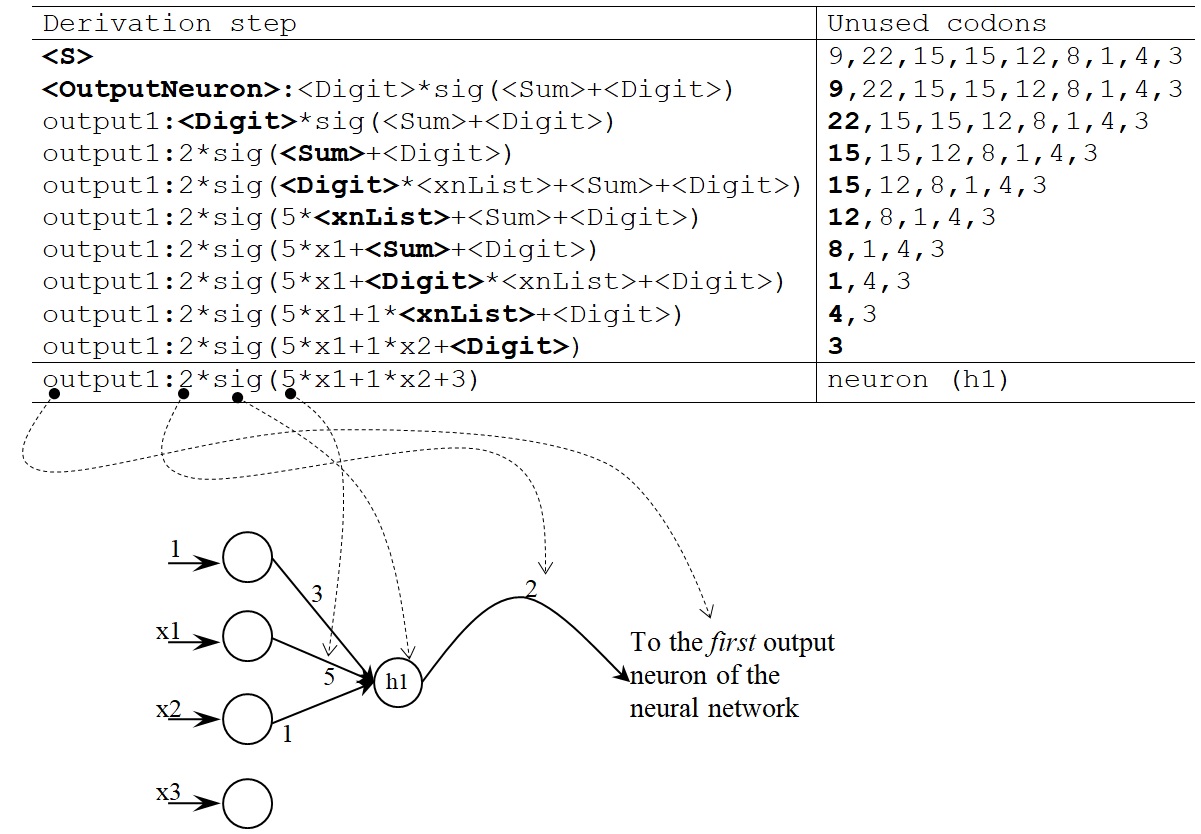}
\caption{Gene to neuron mapping, illustrating  how the first gene of Figure \ref{fig:individual} is mapped to its corresponding neuron (\textit{h1}).}
\label{fig:mapping}
\end{figure}

\textit{Neuron mapping example (See Figure \ref{fig:individual}):} Figure \ref{fig:cfg3} illustrates a simplified version of the proposed CFG in Figure \ref{fig:cfg2} which assigns a single-digit integer to each connection, substitutes the non-terminal $\langle\hspace{0pt}OutputConns\rangle\hspace{0pt}$ with $\langle\hspace{0pt}OutputNeuron\rangle\hspace{0pt}:\langle\hspace{0pt}Digit\rangle\hspace{0pt}$ assuming that each hidden neurons connects to one output neuron, and considers 3 output neurons in the networks.
The simplicity allows  generating a neuron in fewer steps of derivations which makes our example more tractable. 
Figure \ref{fig:mapping} depicts the mapping process steps in rows where the last row shows the output of the process: ``$1:2*sig(5*x_1 + 1*x_2 + 3*1)$" and the first row contains the start symbol ($\langle\hspace{0pt}S\rangle\hspace{0pt}$).
In each derivation step, the left-most non-terminal in the generated expression as well as the current codon are shown in boldface. 
For instance, in the second step, the active non-terminal is $\langle\hspace{0pt}OutputNeuron\rangle\hspace{0pt}$ and the active codon is 9.
Non-terminal $\langle\hspace{0pt}OutputNeuron\rangle\hspace{0pt}$ has 3 production rules ({\it output1}, {\it output2}, and {\it output3}).
Thus, the result of $9 \mod 3 =0$ determines that the first terminal in the production list (\textit{output1}) substitutes the non-terminal.
After each step, we remove the used codon from the presented list.
\fussy 

\subsubsection{Assemble Neurons in a Network}
\label{sssec:assemble}
\begin{figure}[h]
\center
\includegraphics[width=0.90\textwidth]{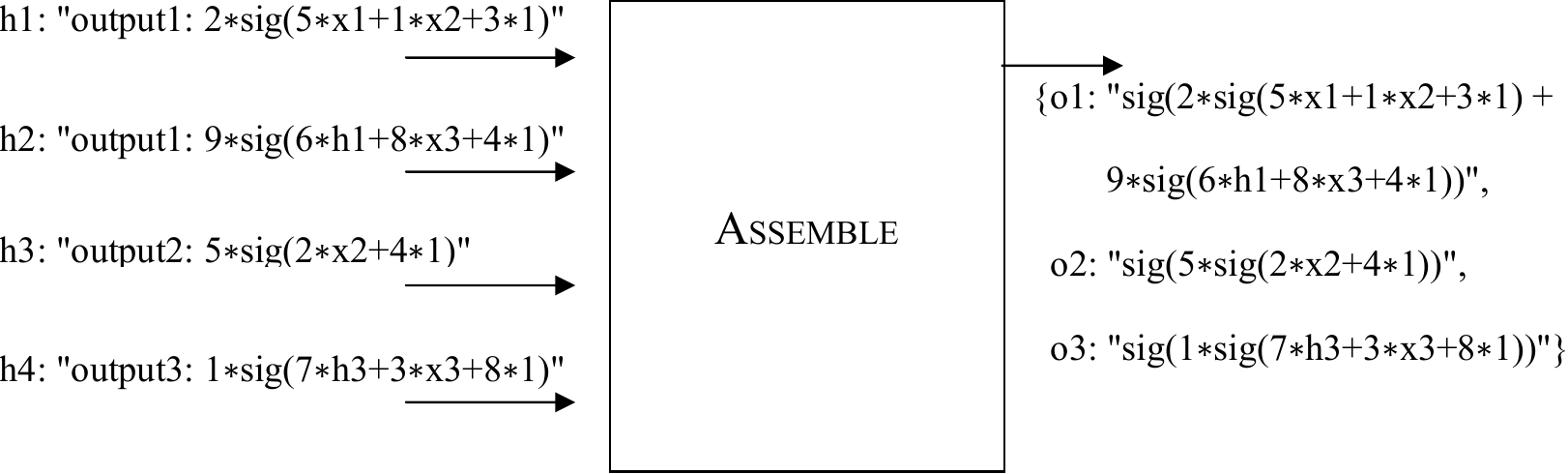}
\caption{{\it h1}, {\it h2}, {\it h3}, and {\it h4} are the neurons depicted in Figure \ref{fig:individual}. Assemble takes {\it h1}, {\it h2}, {\it h3}, and {\it h4} as input and returns three strings representing a neural network with these four hidden neurons and three output neurons.}
\label{fig:assemble}
\end{figure}

\Call{Assemble}{} is the procedure that puts together the neurons leading to a neural network with \textit{c} output neurons (addressing RQ2 in Section \ref{sec:problemstatement}).
The input of Algorithm 2 includes a set of neuron strings in this format: $``\langle\hspace{0pt}OutputConns\rangle\hspace{0pt}$ * sig($\langle\hspace{0pt}Sum\rangle\hspace{0pt}$ + $\langle\hspace{0pt}Number\rangle\hspace{0pt})"$ where the product is a neural network represented as a list of mathematical expressions corresponding to output neurons.
After defining the generated network as an empty list in Line 2, the \textbf{for} loop of Lines 3-13 iterates \textit{c} times to generate an string for each output neuron.
Line 4 initializes the string for $i$-th output neuron with a sigmoid function (\textit{output}\textsubscript{i}). 
Lines 5-10 iterate  through all the hidden neurons and append the string of the neurons that direct to the $i$-th output neuron to \textit{output}\textsubscript{i}.  
Line 6 checks whether or not the string of \textit{neuron} connects to the $i$-th output; 
if so, then Line 7 removes the extra part of the neuron string that represents the output neuron index (i.e., $``\langle\hspace{0pt}OutputNeuron\rangle\hspace{0pt}:"$), and Line 8 appends the modified neuron string to the \textit{output}\textsubscript{i} with an arithmetic plus operation.
For example, the neuron string $``(output5:2),(output3:1)*sig(5*x_1 + 1*x_2 + 3*1)"$ connects to $5$-th output neuron by weight 2 and connects to 3th output neuron by weight 1. 
Then, we should append the modified string $``2*sig(5*x_1 + 1*x_2 + 3*1)"$ to the \textit{output}\textsubscript{5} variable and likewise $``1*sig(5*x_1 + 1*x_2 + 3*1)"$ to the \textit{output}\textsubscript{3} variable.
When the \textbf{for all} loop of Line 5-10 terminates,  Line 11 puts the bracket at the end of output sigmoid function that Line 4 defines. 
Then, Line 12 adds the generated output neuron to the \textit{network} list.
Finally, the algorithm returns the list of \textit{c} generated strings as the produced neural network (Line 14) with $c$ outputs.
For example, Figure \ref{fig:assemble} illustrates how the neurons of Figure \ref{fig:individual} assemble in a neural network.

\begin{algorithm}
\label{alg:alg2}
\caption{Procedure that assembles neurons in a neural network}
\begin{algorithmic}[1]
\Procedure {Assemble}{$neurons$, $c$} 
\item[] \(\triangleright\) $neurons:$ a set of neuron strings in this form: \small
``$\langle\hspace{0pt}OutputConns\rangle\hspace{0pt}$ * sig($\langle\hspace{0pt}Sum\rangle\hspace{0pt}$ + $\langle\hspace{0pt}Number\rangle\hspace{0pt}) $''.
\item[] \(\triangleright\) $c:$ the number of output neurons 
\normalsize
\item[]
\State $network\gets []$
\For {$i:=1$ \textbf{to} $c$}
\State $output\textsubscript{i} \leftarrow ``sig("$
\ForAll {$neuron \in neurons$}
\If {$neuron$ directs to $i$-th output neuron}
\State $neuron' \leftarrow$ remove extra parts (i.e., neuron indices and output weights except for the weight related to $i$-th output) from $neuron$ string
\State $output\textsubscript{i} \leftarrow \Call{ConCat}{output\textsubscript{i}, ``+", neuron'}$
\EndIf
\EndFor
\State $output\textsubscript{i} \leftarrow \Call{ConCat}{output\textsubscript{i}, ``)"}$
\State $network \gets network \cup output\textsubscript{i}$
\EndFor

\State \textbf{return} $network$
\EndProcedure
\end{algorithmic}
\end{algorithm} 

\subsubsection{Neural Network Execution} 
This section describes how a generated ANN predicts examples' class. 
For a {\it c}-class classification problem ({\it c}$>$2), the generated ANN contains {\it c} output neurons with soft-max activation function to generate the posterior probabilities for the {\it c} classes.
If we consider soft-max takes $s_k(x), (1\leq k\leq c)$ as the {\it k}-th output neuron's value for an input pattern X, 
the soft-max outputs the probability of each class as follows:
\begin{equation}
\label{eq:softmax}
o_k(X) = \frac{e^{s_k(X)}}{\sum\limits_{k=1}^{c} e^{s_k(X)}}
\end{equation}
\noindent That is, $o_k(x)$ is the likelihood that the input pattern X is in class {\it k}. However, for binary-class datasets, a single output neuron running a sigmoid activation function is enough to produce the posterior probabilities; when the neuron outputs $s_1 \in [0, 1]$, the occurrence probability of one of the classes having input pattern X would be $s_1$ and the other class gets probability=$\left(1-s_1\right)$.
\\

\subsection{Initialization}
The initialization step, as the first step of MGE, generates $\mu$ individuals randomly to fill the population. 
It runs the following three steps to initialize each individual's genotype:

\begin{enumerate}
\item Determine the number of genes ($\sigma$) randomly in a specified range that user sets (e.g., in this paper: $\sigma \in [2, 10]$ for binary and ternary datasets and $\sigma \in [30, 40]$ for others),
\item Create $\sigma$ genes with pre-defined length ({\it m}),
\item Fill the genes with random integers in range [0, 255].
\end{enumerate}

The parameter {\it m} constrains the complexity of neuron. 
While the evolutionary process can specify {\it m}, for simplicity we set {\it m} to {\it 100}.

\subsection{Crossover and Mutation}
Defining crossover and mutation plays a significant role in the performance and even the success of an evolutionary algorithm.

\begin{figure}[b]
\center
\includegraphics[width=0.8\textwidth]{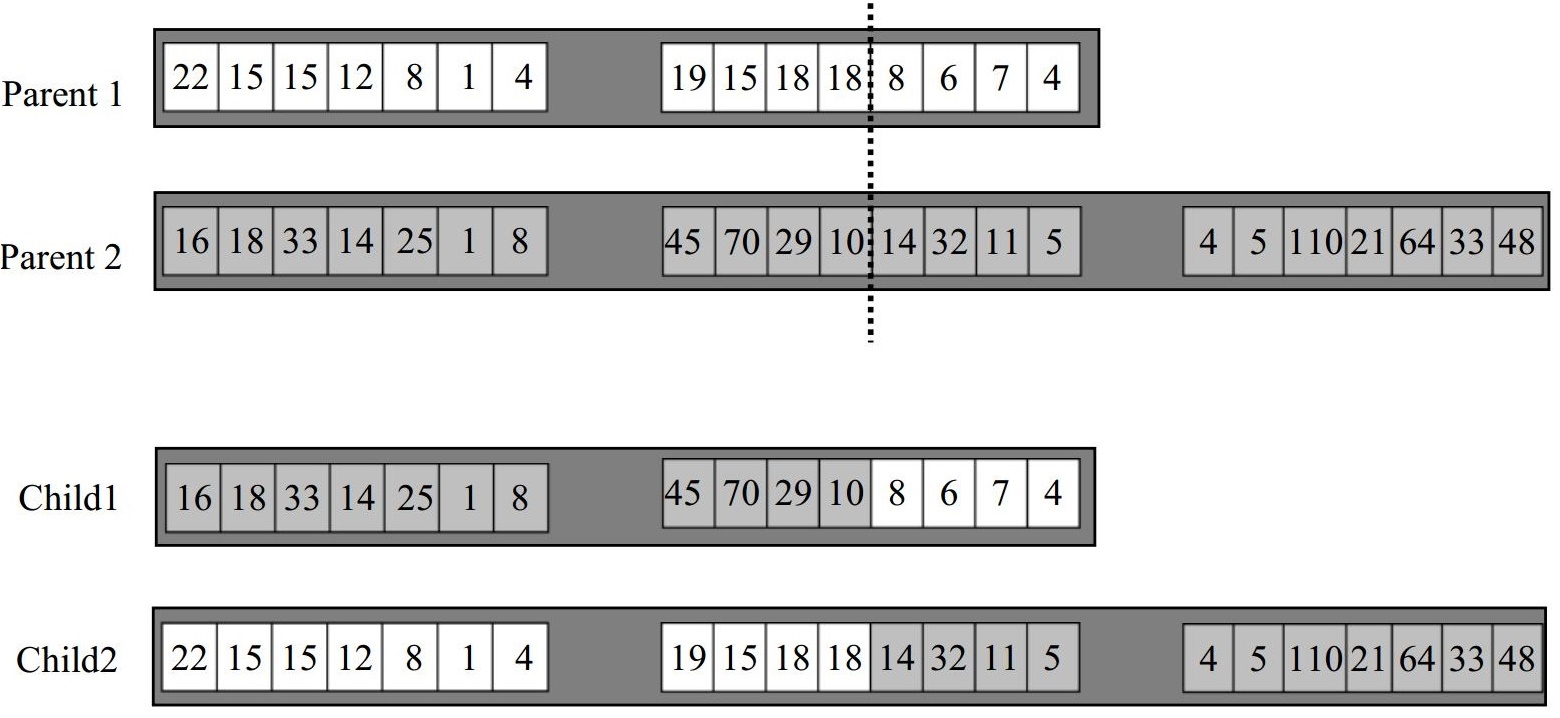}
\caption{The proposed crossover example. The parents' genes are aligned before performing one-point crossover.}
\label{fig:crossover}
\end{figure}

\noindent \textbf{Crossover:} The crossover operator usually serves as the main exploration operator.
It produces two offspring by recombining the genotypes of two parents.
While GE adopts the one-point crossover for the variable length genotypes, we use the one-point crossover in a simpler way.
The crossover between two parents runs in three steps: (1) two parents' genotype are aligned on the left, while their length may differ, (2) a crossover point is selected at random based on smaller genotype, and then (3) the segments on the left hand side of genotype are swapped.
Figure \ref{fig:crossover} depicts an example.

\noindent \textbf{Mutation:} The mutation operator executes on all/some of the produced offspring to make random changes with relatively small chances.
For each offspring, we draw a mutation probability, $P_m$, randomly from the set \textit{\{0.001, 0.002, 0.003, 0.01\}}.
We propose two mutation operations for our new representation. 
The ``gene addition/deletion'' mutation adds or deletes one gene to/from the genotype with probability equal to parameter $P_m$. 
The mutation empowers the algorithm to determine the right number of hidden neurons required to achieve the desired accuracy and generalization ability. 
Mutation through ``codon perturbation'' replace each codon by a random integer drawn from the range [0, 255] with a probability of $P_m/2$.
This latter is the standard GE's mutation operator.

\subsection{Selection Mechanisms}
The parent selection mechanism selects $\mu$ parents randomly with replacement, from the current population with size $\mu$.
To select each parent, we pick {\it r} members at random with replacement from the population, and then we establish a tournament among them according to their fitness. A copy of the winner moves to the parents population.
The parents undergo crossover and mutation to generate $\mu$ offspring. 
Furthermore, the simple replacement strategy with elitism is adopted to form the next generation using both the current population and their offspring.
$P_{elite}$ percentage of the new population is filled with the best individuals of the current population, and 
then fills the rest by the best individuals (according to their fitness) in offspring population.

\subsection{Fitness Function}
\label{sec:fitness}
We use mean of cross-entropy loss to evaluate the ANNs against a classification task represented by a set of training examples. 
If we consider {\it n} to denote the number of patterns in the training data, the following formula shows how the cross-entropy averaged over {\it n} patterns is calculated for individual {\it I}:
\begin{equation}
\label{eq:entropy}
loss (I)= - \frac{1}{n}\sum\limits_{z=1}^{n} \sum\limits_{k=1}^{c} \left(y_{zk} \log \mleft(p_{zk}^{I}\mright)\right)
\end{equation}
where $y_{zk}$ is the binary indicator of desired outputs; it is 1 when {\it k} is the correct class for input pattern {\it z} and 0 otherwise. $p_{zk}^{I}$ denotes the probability that network {\it I} predicts class {\it k} for pattern {\it z}. 
Individuals with lower loss are fitter.

\section{Defining Modularity}
\label{sec:module}

\begin{figure}[h]
\center
\includegraphics[width=0.7\textwidth]{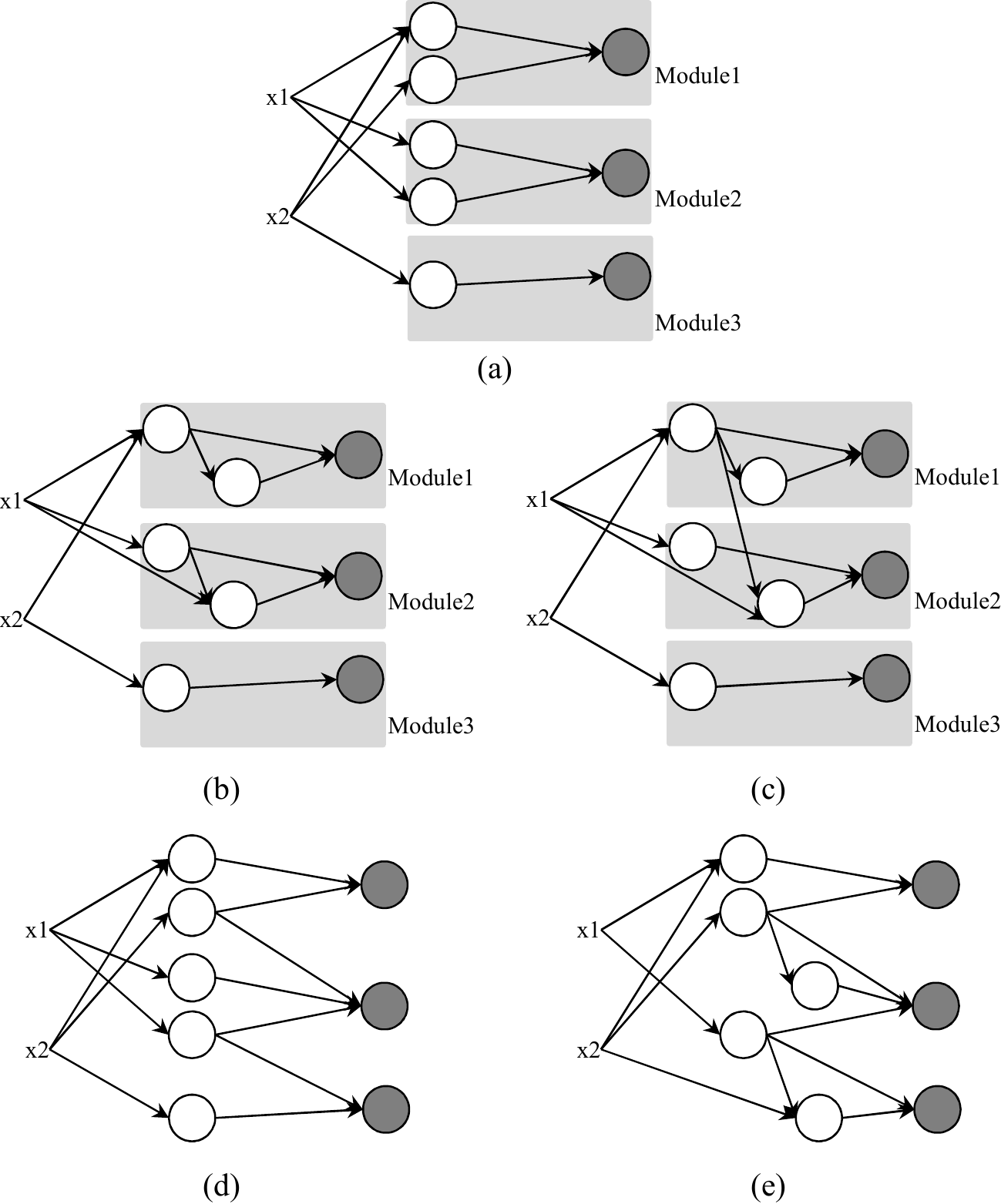}
\caption{Alternatives for defining a module. 
(a) single-layer modules that are highly cohesive with no coupling to other modules, called MGE; (b) multi-layer modules that are highly cohesive with no coupling to other modules, called \textalpha-MGE; (c) multi-layer modules with a single output neuron that are highly cohesive and have high coupling to other modules, called \textbeta-MGE; (d) monolithic and single-layer network, called \texteta-MGE, and (e) monolithic and multi-layer network, called \textmu-MGE. }
\label{fig:versions}
\end{figure}

In this section, we study the question of what a module is and how it should be structured internally and externally (i.e., RQ4, stated in Section \ref{sec:problemstatement}). We postulate an abstract notion of a module, where a {\it module} is a collection of neurons. As such, we pose the following question: How should the  neurons in a module be connected to each other and to the neurons outside the module? To address this question, we follow the principles of Software Engineering and use the two properties of coupling and cohesion for the modules. Coupling refers to how a module connects with other modules and cohesion relates with the internal connectivity/structure of a module and its reliance on its own neurons. To simplify our presentation, we consider 5 different orientations, depicted in Figure \ref{fig:versions}, with a few neurons. 

An ideal scenario occurs when we have a high degree of cohesion and almost no coupling (Figure \ref{fig:versions}-(a) and (b)). That is, a module has no connection to other modules and its neurons are connected to each other or its output neuron. Under this definition, we can have two types of modules, namely single layer (Figure \ref{fig:versions}-(a)) and multi-layer (Figure \ref{fig:versions}-(b)). 
The first MGE (Figure \ref{fig:versions}-(a)) limits the solution space to single-layered neural networks and it allows each hidden neuron to connect to one output neuron.
The hidden neurons connected to an output neuron are considered as a module. Modules in this case have high cohesion and no coupling. The second variant of MGE (Figure \ref{fig:versions}-(b)), called \textalpha-MGE, again limits the solution space to modules  where each hidden neuron only connects to one output neuron, but now internal neurons can be configured in a multi-layer fashion.
 That is, multi-layer modules with high cohesion and no coupling. The third variant of MGE (Figure \ref{fig:versions}-(c)), called \textbeta-MGE, captures modules that allow inter-module and intra-module connections; i.e., high cohesion and  high coupling. 
In the fourth variant of MGE (Figure \ref{fig:versions}-(d)), denoted \texteta-MGE, the networks are single-layered and monolithic. That is, connection between hidden neurons is weak  and a hidden neuron may connect to multiple output neurons. 
The last variant of 
MGE (Figure \ref{fig:versions}-(e)), is multi-layered and monolithic where the hidden neurons may connect to multiple output neurons and the connection between pairs of hidden neurons is possible. 
The MGE and \textmu-MGE respectively impose the highest and lowest restrictions on the search space.
\textalpha-MGE and \textbeta-MGE examine how MGE behaves in the presence of modularity but with multi-layered topology.
\texteta-MGE and \textmu-MGE test the ablation of modularity on the classification performance of MGE in respectively single-layered and multi-layered network topologies. 

\subsection{Enforcing Modularity} 

The presented method that uses the mapping process of Algorithm 1 and the CFG of Figure \ref{fig:cfg2} enforces the least restriction on the generated neural network topology as it enables every hidden neuron to connect to multiple output neurons and also connect to other hidden neurons. 
As a result, the generated neural networks lay in the group of monolithic multi-layered neural networks or \texteta-MGE.
To generate single-layered monolithic neural networks, or \textmu-MGE, we simply should comment Line 11 of Algorithm 1.
For modular variants (MGE, \textalpha-MGE, and \textbeta-MGE), where each hidden neuron is limited to connect to only one output neuron, we should modify the first rule of the CFG as follows: $``(\langle\hspace{0pt}OutputNeuron\rangle\hspace{0pt}:\langle\hspace{0pt}Number\rangle\hspace{0pt})$ * sig($\langle\hspace{0pt}Sum\rangle\hspace{0pt}$ + $\langle\hspace{0pt}Number\rangle\hspace{0pt})"$.
Figure \ref{fig:individual} depicts an example of modular neural networks, where we show its generating grammar in Figure \ref{fig:cfg3}.  
As Figure \ref{fig:individual} shows, Algorithm 1 generates the networks that lay in the group of \textbeta-MGE, where the hidden neurons may connect to each other even if they are in different modules.
To generate the single-layered modular neural networks (referred as MGE products), we simply should comment Line 11 of Algorithm 1 in addition to changing the first rule of the used CFG as described.
However, for \textalpha-MGE networks, Algorithm 1 should change in another way: (1) after initializing {\it grammar'} and {\it neuronsCount} variables, the algorithm needs to group the genes into {\it c} different lists; the modulo of a gene's first codon to {\it c} determines which group the gene will go, (2) then, the algorithm maps the neurons of a group, and (3) resets the grammar before mapping the neurons of next group.

Moreover, the simplicity of the generated neural networks and algorithm's efficiency are important aspects that we consider.
More generalization ability, faster execution during runtime and simplicity of hardware implementations (for special applications) are three potential outcomes of simpler neural networks.

\section{Experimental Evaluation of the  Definitions of Modularity}
\label{sec:exper1}

This section experimentally evaluates the impact of each variant of MGE defined in Section \ref{sec:module} on training performance and test accuracy. We use 10 well-known classification benchmarks that are used by various previous methods: Wisconsin Breast Cancer Detection (WDBC), Ionosphere, Sonar, Credit German, Heart Disease, Wine and Letter taken from UCI repository \citep{UCI}, Flame \citep{Flame}, Handwritten Digits dataset \footnote{Available at https://scikit-learn.org/stable/datasets/}, and MNIST dataset \citep{Hinton2006, Bengio2006} \footnote{Available at http://yann.lecun.com/exdb/mnist/}.
Table \ref{tab:datasets} presents the number of features, classes and instances  of these datasets.  We initially performed some preliminary experiments on a random set of aforementioned datasets to determine the appropriate values of the parameters involved in the proposed method. 
Table \ref{tab:params} presents the achieved parameter settings that we use in our experiments. 
For implementation, we extended Grammatical Evolution in Java (GEVA) \citep{GEVA} library.


\begin{table*}
\caption{The datasets' properties. {\it Iono.} is the abbreviation for the Ionosphere dataset.}
\begin{tabular}{ p{0.09\textwidth}| p{0.053\textwidth} p{0.06\textwidth} p{0.053\textwidth} p{0.053\textwidth} p{0.053\textwidth} p{0.053\textwidth} p{0.053\textwidth} p{0.053\textwidth} p{0.053\textwidth} p{0.071\textwidth}} \hline 
& Flame & WDBC & Iono. & Sonar & Credit & Heart & Wine & Digits & Letter & MNIST\\ \hline 
Features & 2 & 30 & 34 & 60 & 20 & 13 & 13 & 64 & 16& 784 \\ \hline 
Classes & 2 & 2 & 2 & 2 & 2 & 2 & 3 & 10& 26 & 10\\ \hline
Instances & 240 & 569 & 351 & 208 & 1000 & 270 & 178 & 1797 & 20000 & 70000\\ \hline 
\end{tabular}
\label{tab:datasets}
\end{table*}

Table \ref{tab:versions} presents the experimental results averaged over 30 independent runs for the evaluation of the variants of MGE (introduced in Section \ref{sec:module}). The reported results include Root Mean Square Error (RMSE) of the ANNs on training data (represented by RMSE\textsubscript{n}), the RMSE on testing data (RMSE\textsubscript{t}) and accuracy of the networks on testing data (Acc.\textsubscript{t}) as well as the number of hidden layers (\#H.L.), hidden neurons (\#Neu.), and the number of input features (\#Fea.). 
The results for the aforementioned criteria are averaged and Table \ref{tab:versions} presents the outcome in this format: average$\pm$standard deviation.
Observe that, for binary class datasets, the definitions of  \texteta-MGE and MGE collapse to the same definition, and the other three definitions \textalpha-MGE, \textbeta-MGE and \textmu-MGE become  identical too. 
As a result, for the first 6 datasets, we report only the results of MGE and \textmu-MGE. 
In these cases, MGE and \textmu-MGE respectively generate  single-layered  and multilayered neural networks where all the neural networks are monolithic.

\begin{table}[h]
\center
\caption{The parameter settings used in the experiments.}
\label{tab:params}
\begin{tabular}{p{2.1in}|p{2.5in}} \hline 
\textbf{MGE Parameter} & \textbf{Value} \\ \hline 
Population size ($\mu$) & 200 \\ 
Max no. of generations & 500 for binary and ternary datasets, and 3000 for others \\ 
Crossover rate (${P}_{c}$) & 0.9 \\ 
Probability of mutation (${P}_{m}$) & $\in$ \{0.001, 0.002, 0.003, 0.01\} \\ 
Tournament size& 7 \\ 
Elite size ($P_{elite}$) & 0.05 \\ 
Gene length ($m$) & 100 \\ 
The range of initial num. of genes& [2, 10] for binary and ternary datasets, and [30, 40] for others \\ 
Wrapping & 0 \\ \hline 
\end{tabular}
\end{table}

\noindent \textbf{Statistical Tests}.\
To see if the distance between the classification accuracy of two algorithms is significant we use a statistical test with a significance level of 0.025. 
The best classification accuracy for each dataset is highlighted in bold, where multiple bold faced accuracies indicate that these accuracies are statistically equal and significantly better than those with non-bold face values. 
Since the classification accuracies of the algorithms do not necessarily follow a normal distribution  \citep{Assunco-gecco}, most of the new articles use a Wilcoxon test that is a non-parametric test \citep{Assunco-gecco, McDonnell, Watts}. 
Accordingly, we perform Wilcoxon tests to assess the differences between classification accuracies.

\begin{table*}[ht!]
\setlength\tabcolsep{1pt}
\caption{Comparisons between MGE and other variants of MGE: \textalpha-MGE, \textbeta-MGE, \texteta-MGE, and \textmu-MGE.\newline }
\label{tab:versions}
\begin{tabular}{p{0.09\textwidth}|p{0.10\textwidth}|p{0.12\textwidth}|p{0.12\textwidth}|p{0.13\textwidth}|p{0.12\textwidth}|p{0.13\textwidth}|p{0.11\textwidth}} \hline
Dataset & Method & RMSE\textsubscript{n} & RMSE\textsubscript{t} & Acc.\textsubscript{t} & \#H.L. & \#Neu. & \#Fea.\\ \hline 
\multirow{2}{*}{Flame} & MGE & 0.10$\pm$0.02 & 0.11$\pm$0.02 & 0.982$\pm$0.01 & 1.00$\pm$0.00& 7.13$\pm$1.63& 2.00$\pm$0.00 \\

& \textmu-MGE & 0.10$\pm$0.03 & 0.12$\pm$0.02 & 0.977$\pm$0.01 & 1.33$\pm$0.46& 6.90$\pm$1.40& 2.00$\pm$0.00 \\ \hline 

\multirow{2}{*}{WDBC} & MGE & 0.10$\pm$0.02 & 0.18$\pm$0.03 & \textbf{0.967$\pm$0.02} & 1.00$\pm$0.00 & 8.00$\pm$1.86 & 13.57$\pm$2.5 \\

& \textmu-MGE & 0.14$\pm$0.03 & 0.20$\pm$0.03 & 0.951$\pm$0.02 & 1.93$\pm$0.57& 8.13$\pm$1.63& 12.50$\pm$3.3\\ \hline 

\multirow{2}{*}{Iono.} & MGE & 0.14$\pm$0.03 & 0.27$\pm$0.03 & 0.917$\pm$0.02 & 1.00$\pm$0.00& 9.53$\pm$1.87& 14.40$\pm$3.7\\ 

& \textmu-MGE & 0.16$\pm$0.02 & 0.28$\pm$0.03 & 0.915$\pm$0.02 & 1.83$\pm$0.73& 9.03$\pm$1.43& 13.46$\pm$3.3 \\ \hline

\multirow{2}{*}{Sonar} & MGE & 0.28$\pm$0.01 & 0.39$\pm$0.02 & \textbf{0.793$\pm$0.03} & 1.00$\pm$0.00& 11.50$\pm$2.59& 17.07$\pm$3.5\\

& \textmu-MGE & 0.27$\pm$0.02 & 0.40$\pm$0.03 & 0.785$\pm$0.04 & 2.03$\pm$0.60& 10.13$\pm$1.75& 15.93$\pm$2.9\\ \hline
 
\multirow{2}{*}{Heart} & MGE & 0.24$\pm$0.02 & 0.36$\pm$0.02 & 0.827$\pm$0.03 & 1.00$\pm$0.00& 7.40$\pm$2.00 & 6.93$\pm$1.23 \\

& \textmu-MGE & 0.27$\pm$0.02 & 0.40$\pm$0.03 & 0.822$\pm$0.03 & 1.63$\pm$0.55& 6.77$\pm$1.45 & 6.47$\pm$1.43 \\ \hline 

\multirow{2}{*}{Credit} & MGE & 0.30$\pm$0.02 & 0.41$\pm$0.02 & \textbf{0.754$\pm$0.02} & 1.00$\pm$0.00& 10.53$\pm$2.91& 15.10$\pm$2.7 \\

& \textmu-MGE & 0.31$\pm$0.02 & 0.44$\pm$0.03 & 0.741$\pm$0.02 & 2.03$\pm$0.71& 11.17$\pm$2.90& 14.13$\pm$3.0 \\ \hline 

\multirow{5}{*}{Wine} & MGE & 0.15$\pm$0.03 & 0.31$\pm$0.02 & \textbf{0.934$\pm$0.04} & 1.00$\pm$0.00& 10.67$\pm$1.27& 8.06$\pm$1.26 \\ 

& \textalpha-MGE & 0.18$\pm$0.02 & 0.33$\pm$0.04 & 0.902$\pm$0.03 & 1.73$\pm$0.77& 11.13$\pm$1.67&7.87$\pm$1.17 \\ 

& \textbeta-MGE & 0.17$\pm$0.03 & 0.34$\pm$0.02 & 0.901$\pm$0.03 & 1.96$\pm$0.71& 10.50$\pm$1.43&7.60$\pm$1.20 \\

& \texteta-MGE & 0.18$\pm$0.04 & 0.35$\pm$0.03 & 0.861$\pm$0.04 & 1.00$\pm$0.00& 10.33$\pm$1.16& 8.16$\pm$1.39 \\ 

& \textmu-MGE & 0.20$\pm$0.05 & 0.37$\pm$0.04 & 0.857$\pm$0.03 & 2.30$\pm$0.74& 11.40$\pm$1.45 & 7.33$\pm$1.32 \\ \hline 

\multirow{5}{*}{Digits} & MGE & 0.18$\pm$0.03 & 0.27$\pm$0.01 & \textbf{0.945$\pm$0.02} & 1.00$\pm$0.00& 55.5$\pm$2.32& 59.80$\pm$3.5 \\

& \textalpha-MGE & 0.19$\pm$0.03 & 0.27$\pm$0.02 & 0.893$\pm$0.03 & 4.10$\pm$1.54& 45.10$\pm$3.91& 51.93$\pm$2.0 \\

& \textbeta-MGE & 0.20$\pm$0.04 & 0.28$\pm$0.02 & 0.890$\pm$0.03 & 3.33$\pm$1.32& 47.17$\pm$3.68& 52.03$\pm$2.7 \\

& \texteta-MGE & 0.19$\pm$0.02 & 0.32$\pm$0.02 & 0.870$\pm$0.01 & 1.00$\pm$0.00& 57.17$\pm$3.50& 58.23$\pm$3.1 \\ 

& \textmu-MGE & 0.22$\pm$0.04 & 0.32$\pm$0.01 & 0.857$\pm$0.03 & 3.87$\pm$0.96& 45.8$\pm$2.96& 50.97$\pm$2.2\\ \hline 

\multirow{5}{*}{Letter} & MGE & 0.55$\pm$0.03 & 0.61$\pm$0.01 & \textbf{0.547$\pm$0.02} & 1.00$\pm$0.00& 74.47$\pm$4.93& 15.03$\pm$1.0 \\

& \textalpha-MGE & 0.59$\pm$0.03 & 0.63$\pm$0.03 & 0.493$\pm$0.05 & 3.97$\pm$0.98& 79.37$\pm$6.35& 15.70$\pm$0.9 \\ 

& \textbeta-MGE & 0.62$\pm$0.04 & 0.65$\pm$0.02 & 0.495$\pm$0.06 & 3.80$\pm$0.65& 75.04$\pm$11.9& 15.20$\pm$1.1 \\ 

& \texteta-MGE & 0.60$\pm$0.02 & 0.73$\pm$0.02 & 0.424$\pm$0.04 & 1.00$\pm$0.00& 75.67$\pm$4.85& 15.23$\pm$1.3 \\

& \textmu-MGE & 0.62$\pm$0.03 & 0.70$\pm$0.02 & 0.409$\pm$0.05 & 4.20$\pm$0.48& 81.00$\pm$8.54& 15.80$\pm$0.4 \\ \hline 

\multirow{5}{*}{MNIST} & MGE & 0.30$\pm$0.02 & 0.35$\pm$0.02 & \textbf{0.893$\pm$0.02} & 1.00$\pm$0.00& 63.93$\pm$4.62& 177.1$\pm$11 \\

& \textalpha-MGE & 0.33$\pm$0.03 & 0.39$\pm$0.02 & 0.819$\pm$0.05 & 4.03$\pm$1.22& 64.17$\pm$5.77& 175.7$\pm$12 \\

& \textbeta-MGE & 0.34$\pm$0.01 & 0.40$\pm$0.02 & 0.776$\pm$0.03 & 4.77$\pm$1.75& 65.13$\pm$7.64& 168.0$\pm$18 \\

& \texteta-MGE & 0.36$\pm$0.02 & 0.41$\pm$0.02 & 0.759$\pm$0.02 & 1.00$\pm$0.00& 64.53$\pm$3.24& 167.8$\pm$17 \\

& \textmu-MGE & 0.35$\pm$0.03 & 0.42$\pm$0.03 & 0.766$\pm$0.04 & 4.80$\pm$1.05 & 66.93$\pm$8.07& 180.6$\pm$17 \\ \hline 
\end{tabular}
\end{table*}

\noindent{\bf Training Performance}.\ 
Table \ref{tab:versions} demonstrates that the RMSE\textsubscript{n} of MGE is less than that of \textmu-MGE in 3 cases of the first 6 datasets, while in other 3 cases they are equal.
For Wine, Digits, Letter, and MNIST datasets, the RMSE\textsubscript{n} of MGE is the lowest in all cases, and the RMSE\textsubscript{n} of modular topologies generated by MGE, \textalpha-MGE, and \textbeta-MGE are lower or equal to that of monolithic topologies (generated by \texteta-MGE and \textmu-MGE) in almost all cases.
In summary, the results demonstrate that single-layered and modular topology of Figure \ref{fig:versions}-(a) increases the training performance.

\noindent{\bf Test Accuracy}.\ 
Table \ref{tab:versions}, column Acc.\textsubscript{t} demonstrates that among the modular variants, single-layered neural networks return statistically better accuracies than multi-layered neural networks for all of the four datasets Wine, Digits, Letter, and MNIST.
If we compare MGE with \textmu-MGE on the first 6 datasets, the single-layered monolithic neural networks statistically return better accuracies than multi-layered monolithic networks in 3 cases.
In the remaining 3 cases, the results of the two variants are statistically equal.
Comparisons of \texteta-MGE and \textmu-MGE on last 4 datasets, demonstrate no superiority of single-layered or multi-layered networks; their accuracies are statistically equal for Wine and Digits, \texteta-MGE's accuracy is better for MNIST, and \textmu-MGE is better for Letter.
Summing up all the  comparisons, single-layered networks win eight cases, single and multi-layered are equal in five cases, and multi-layered networks outperform the rest in one case.
Moreover, the results of multi-class datasets demonstrate that modular topology of Figure \ref{fig:versions}-(a)  significantly improves the neural networks' learning and generalization. 

\noindent{\bf Network Structure}.\ 
MGE's generated neural networks have more neurons if they are single-layered for binary and ternary datasets, while for other datasets the networks generated by all variants of MGE are in the same ballpark.  
However, single-layered neural networks of MGE provide a specific form of simplicity that may compensate the extra neurons they have.
Considering RMSE\textsubscript{n}, RMSE\textsubscript{t}, and Acc.\textsubscript{t}, we conclude that {\em MGE in Figure \ref{fig:versions}-(a) is the most favorable variant}. That is, the notion of a single-layered topology with no coupling to other modules increases the training performance and test accuracy of neural networks generated by MGE. As such, hereafter, we use the terms `MGE' and `proposed modularity' interchangeably. 
The proposed modularity induces an special form of sparsity to the neural networks where {\it each hidden neuron connects to a single output neuron which means a hidden neuron may activate only one output neuron.}
Additionally, the GEs including MGE generate sparse neural networks because of their bias towards simpler solutions. For example, our experiments demonstrate that other GEs generate neural networks up to 90\% of sparsity while those of MGE are close to 80\% sparse for Ionosphere dataset. 
As another interesting feature of GEs, the user is able to define other specific forms of sparsity or tune the sparsity rate of the networks through the input grammar.
The sparsity is biologically plausible \citep{Glorot2011, Attwell2001}; in a human brain only a few fraction of the neurons (about 1\% to 4\%) are activated at same time \citep{Lennie2003}, because a neuron connects to an small number of other neurons (i.e., 2000  out of 86 billion on average) \citep{Frederico2009}. 
From computational neuroscience point of view, the sparsity is an appealing regularization technique that provides more efficient computation and hardware acceleration and the possibility to model larger problems \citep{Glorot2011, Alford2018}.

\section{Experimental Validation of MGE}
\label{sec:exper2}


This section reports the results of our experimental evaluation of MGE (i.e., definition of modularity based on Figure \ref{fig:versions}-(a)) with respect to a few state-of-the-art methods. We compare MGE with relevant state-of-the-art GE-based methods, namely GE \citep{tsoulos2008}, SGE \citep{Assuncao2017}, DSGE \citep{Assunco-gecco}, NEAT \citep{Stanley2002}, and OVA-NEAT \citep{McDonnell}.
OVA-NEAT is an ensemble method that decomposes a \textit{c}-class classification problem into \textit{c} binary class problems,
 and then generates a neural networks for each binary classification problem using NEAT.
The aggregation of the  \textit{c} outputs of neural networks  determines which class is the output of the ensemble neural network.

\noindent{\bf Experiment Setup}.\
For partitioning the datasets to train and test sets, we first shuffle the patterns. 
Then, for the first 7 datasets, we consider 70\% of the patterns as the training data, and the remaining patterns as the test set.
However, for Digits and Letter datasets we follow \citep{McDonnell} and use respectively 90\% of patterns as training data.
We use the MNIST data in a standard partitioned format (60000 patterns as training data and 10000 as testing data). 
The proposed method  uses the training data to calculate the fitness of neural networks during the evolution process as described in Section \ref{sec:fitness} where the test set is kept away from the process, and we use it to assess the fittest solution generated by our  algorithm. 
To overcome the randomness effect of the evolutionary algorithms, we run the process of data partitioning and algorithm execution, independently for 30 times on binary and ternary datasets. While \cite{McDonnell} report the results of 10 independent runs on Digits, Letter, and MNIST, we run MGE for 10 times on these datasets.
Our experimental results present the average of the outputs of these executions.


\subsection{Representation Analysis}
\label{sec:repanalysis}

This section reports on the impact of MGE on the number of invalid individuals, locality of the evolution process, and scalability of the generated networks. 


\noindent \textbf{Invalidity:} In other GE-based mappings, an individual is valid if its mapping terminates in a specific number of derivation steps, otherwise it is invalid even if it leads to many valid components  during its mapping process. 
Invalid individuals get the worst fitness value and will vanish from the population soon while they may include valuable information.
In contrast, our method redefines invalidity by  decoding each gene to a neuron independently, and returning a valid individual even if only one of the genes is mapped successfully.
Our experiments on all of the datasets demonstrate that the occurrence rate of invalid individuals in the population is 0.0005 (i.e., 50 out of 100000 evaluations), while it is more than 0.03 on average for GE.
SGE and DSGE address this problem completely, at the expense of a significant increase in genotype length and extra repair mechanism which repairs the genotype to ensure that the perturbed genotype is mapped successfully to a valid neural network.

\noindent \textbf{Locality:} In GE, SGE, and DSGE a small change to the genotype of a good solution (e.g., a codon perturbation) is likely to ruin it.
Due to  one-to-one correspondence of genes to neurons, MGE raises the locality strength by restricting the codon perturbations to the neurons' domain. 
The proposed modular genotype enables a controlled addition/deletion of up to {\it g} genes (or neurons) to/from a neural network through the mutation operation. 
In this work, we consider {\it g=1}. 

To compare the locality strength in MGE with other algorithms, we design an experiment where for each algorithm we first generate 8 populations for 8 different network sizes (networks with \{2,3, ..., 9\} neurons) where each population contains 5000 random neural networks having a specific neuron count.
Thus, we have 40000 randomly generated individuals for each algorithm.
Then, we apply two mutations on each solution's genotype {\it I} to form two new offspring {\it I'} and {\it I''} with respectively 1 and 2 Hamming bits difference from {\it I}.
Finally, we calculate the phenotype distance between each solution and its mutants using tree edit distance \citep{Pawlik}; lower distance means higher locality.
Tree edit distance is the minimum number of node edit operations (insert, delete, and rename) that transform one tree into another. 
To use the criterion, we convert the phenotypes (neural networks) to their corresponding evaluation trees.
Table \ref{tab:locality} details the network distances averaged on 40000 generated individuals.
The results indicate that MGE is the most promising method to improve locality strength.
Notice that, we use a single grammar for all the methods: the number of features is 30, the number of digits in connection weights is 3 (in this format \#.\#\#), and there is only one hidden layer in the networks.

\begin{table*}
\caption{Tree edit distance between two neural networks when their genotypes are 1 or 2 Hamming bits different. The results are averages and standard deviations (denoted in avg.\ensuremath{\pm}std. format) of 40000 random individuals.}
\center
\begin{tabular}{ p{0.10\textwidth} P{0.2\textwidth} P{0.2\textwidth} } \hline 
Method & 1 bit distance & 2 bits distance\\ \hline 
GE &6.84\ensuremath{\pm}12.21 & 8.48\ensuremath{\pm}17.13 \\ 
SGE & 6.91\ensuremath{\pm}13.54 & 8.30\ensuremath{\pm}15.01 \\ 
DSGE & 7.85\ensuremath{\pm}12.15 & 9.37\ensuremath{\pm}13.71 \\ 
MGE & 1.24\ensuremath{\pm}3.31 & 1.44\ensuremath{\pm}3.2 \\ \hline 
\end{tabular}
\label{tab:locality}
\end{table*} 

\cite{Bartoli} recommend the {\it static} context where the representation analyses apply on a random initialized population instead of on the populations during an evolutionary process. 
The  dynamics provided by search and selection operations during the evolutionary process push the population distribution towards a pre-specified objective function. The analyses of representation properties (i.e., invalidity, locality, and scalability) in such a {\it dynamic} context are dependent upon the objective function \citep{Medvet2017}.

They draw 10000 pairs randomly of 10000 random individuals where the individuals may be far from each other in genotype space.
However, we work a bit different from their method.
We believe that, the locality strength is important and necessary for the exploitation mechanism in an evolutionary algorithm where the perturbations are rare. 
Thus, it is better to examine the locality when the genotypic distance is small, instead of randomly picking pairs.
Moreover, we generate random neural networks with different neuron counts, because every representation has an intrinsic bias towards some specific network sizes that might affect  the results of analyses.

\begin{figure}[h]
\center
\includegraphics[width=0.85\textwidth]{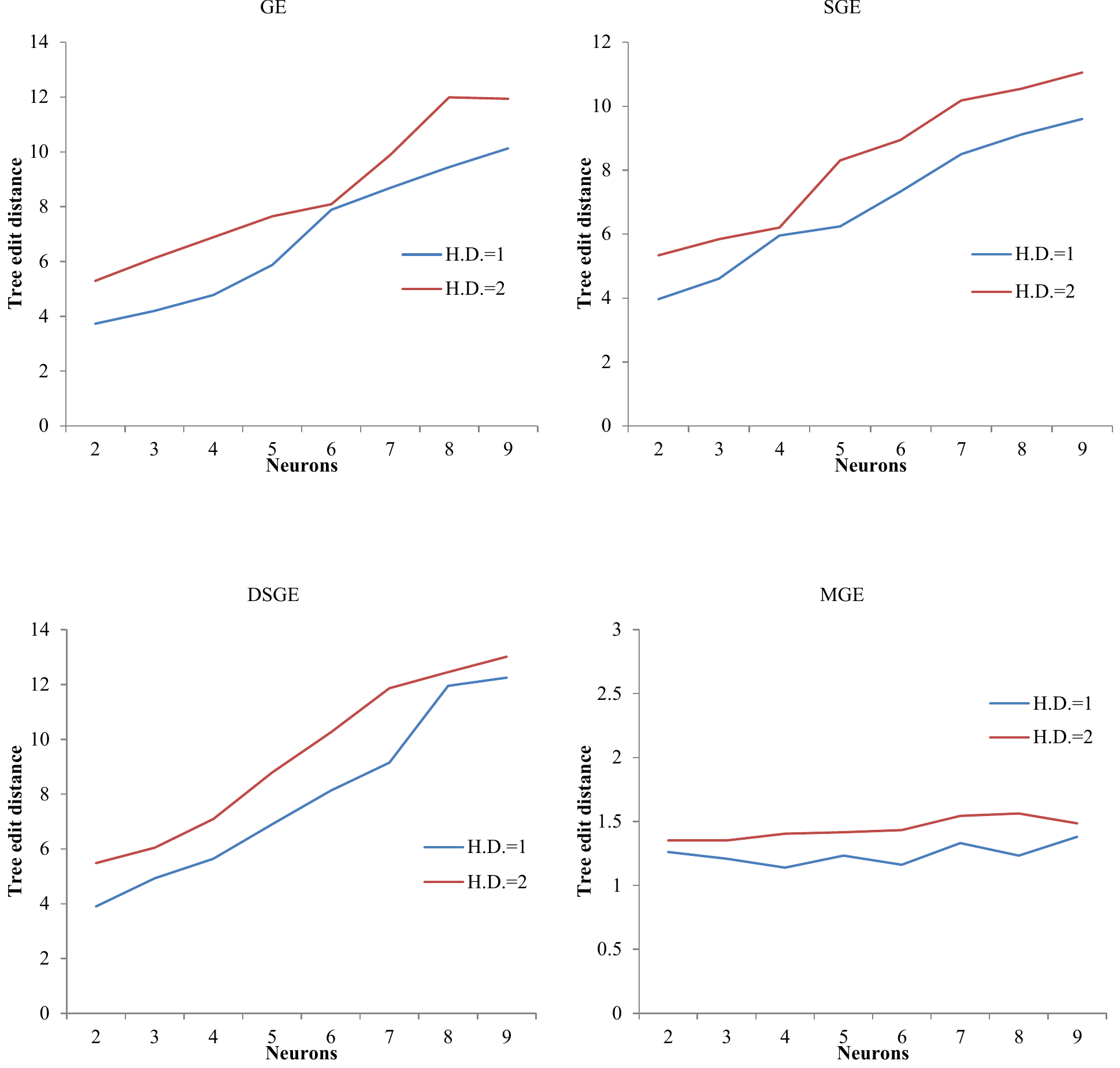}
\caption{The effect of network size (calculated by neuron count) on the locality strength. A curve depicts the phenotype distance caused by 1 or 2 bits of Hamming perturbations on genotypes (denoted by H.M.=1 or H.M.=2) on networks with different neuron counts. Mutation in GE, SGE, and DSGE changes the networks with bigger step sizes when the size of solution increases, while in MGE the step size remains almost fixed.}
\label{fig:locality}
\end{figure}

Additionally, Figure \ref{fig:locality} reveals that when the number of neurons increases, the locality decreases almost linearly for GE, SGE, and DSGE; while it is fairly constant for MGE.
That means, GE, SGE, and DSGE may destroy the bigger solutions with more probability than smaller ones. 

\noindent \textbf{Scalability:} As we mentioned in Section \ref{sec:scalability}, other GE-based representations have bias towards smaller  solutions leading to lower degree of scalability. 
The proposed representation of this paper divides a complex solution's encoding to several encodings of its simpler components.
As such, MGE scales to big problems through adding more components directly to the genotype. 
This approach has other applications such as automatic programming where we divide a program to many reusable functions, each being codified in a distinct gene. 
Indeed, this approach seems to be highly interesting in evolution of deep neural networks where simpler networks are reused and repeated frequently in a deep network \citep{Miikkulainen17}. 

\begin{figure}
\center
\includegraphics[width=0.5\textwidth]{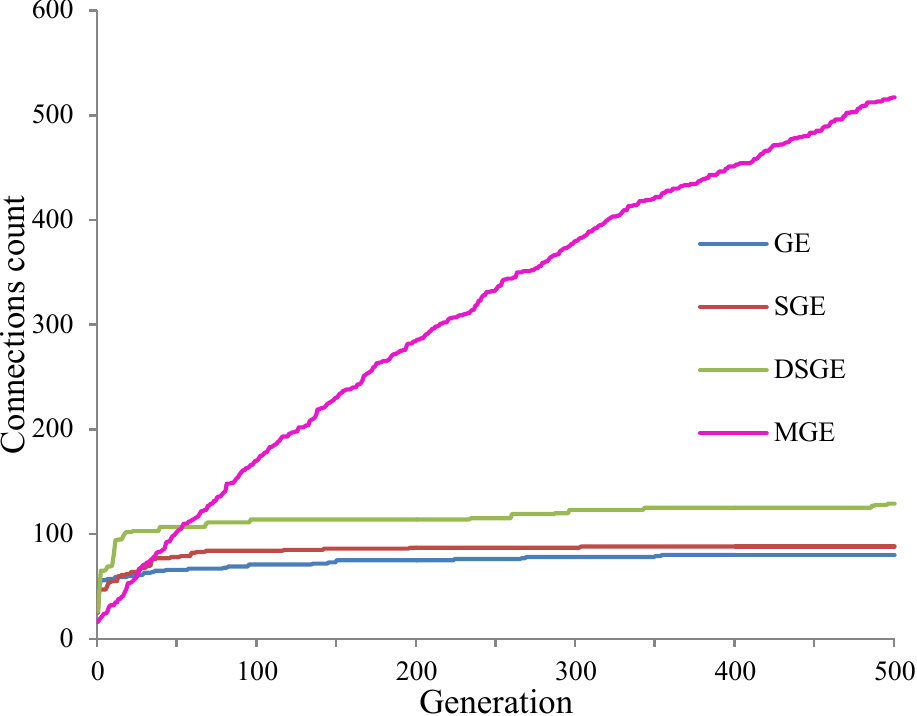}
\caption{Scalability of MGE vs. GE, SGE and DSGE. Here, the objective of the algorithms is to produce networks with more connections.}
\label{fig:connectioncount}
\end{figure}

We design an experiment to compare the scalability of MGE with GE \citep{tsoulos2008}, SGE \citep{Assuncao2017}, and DSGE \citep{Assunco-gecco}.
To this end, we change the objective functions of the algorithms to maximize the \textit{connections count} of the synthesized networks regardless of their classification error. 
Figure \ref{fig:connectioncount} depicts the average connections count of neural networks evolved by the algorithms during 500 generations. 
The results are averaged over 30 independent runs, while all of the conditions that may affect the results are considered for a fair comparison:
(1) we set identical crossover and mutation probability among the algorithms;
(2) we initialize identical genotype length for GE, DSGE and MGE, while setting genotype length of SGE to a far bigger value, as the length in SGE remains fixed during the evolution, and
(3) we introduce an extra mutation for GE \citep{tsoulos2008} that can  change the length of a genotype through the mutation that adds/deletes some codons to/from the genotype,  whereas DSGE may lengthen a genotype through its repair mechanism.
Figure \ref{fig:connectioncount} illustrates the idea that scalability issue is severe in GE methods and demonstrates that MGE addresses the problem successfully. 

\subsection{Comparisons}
\label{sec:comparisons}
To assess the desired properties, we design two sets of experiments to compare the MGE's outputs with other methods.
We also compare the neural networks generated by the algorithms (the fittest individual in the last generation) in terms of their performance on training data, testing data and their complexity.
Table \ref{tab:comp} compares MGE with other algorithms (GE, SGE, DSGE, NEAT, and OVA-NEAT) on these criteria: Root Mean Square Error (RMSE) of the ANNs on training data (represented by RMSE\textsubscript{n}), the RMSE on testing data (RMSE\textsubscript{t}) and accuracy of the networks on testing data (Acc.\textsubscript{t}) as well as the number of hidden layers (\#H.L.), hidden neurons (\#Neu.), and the number of input features (\#Fea.) (for NEAT and OVA-NEAT, we did not have access to complete information).
The results of NEAT and OVA-NEAT for the last 3 datasets (Digits, Letter, and MNIST) are obtained from \cite{McDonnell} while others are obtained from our experiments. 
We re-implemented GE \citep{tsoulos2008}, and revised the code of SGE \citep{Assuncao2017}, and DSGE \citep{Assunco-gecco} taken from {\it Github} account of one of the authors (Louren\c{c}o, N.) \footnote{https://github.com/nunolourenco?tab=repositories}.
For the experiments of GE variants, we use parameter settings that GE \citep{tsoulos2008}, SGE \citep{Assuncao2017}, and DSGE \citep{Assunco-gecco} report. 
Following the above experiment setup (Section \ref{sec:exper2}), every experiment over a dataset in the first 7 datasets generates 30 neural networks, while an experiment over a dataset of the rest produces 10 neural networks. The aforementioned statistics are obtained from these produced neural networks.

Again, we perform statistical tests to see if a distance between the classification accuracy of two algorithms is significant or not. 
We use a Wilcoxon test for the binary and ternary datasets, where we have the accuracy values of entire methods for all runs. 
However, we have only averages and standard deviations of accuracy of peer competitors for Digits, Letter, and MNIST. 
Thus, in the  case of these datasets, we assume that the values follow a normal distribution and perform t-tests with 9 degrees of freedom.

\begin{table*}
\setlength\tabcolsep{1pt}
\caption{Comparisons between MGE and other methods: GE \citep{tsoulos2008}, SGE \citep{Assuncao2017}, DSGE \citep{Assunco-gecco}, NEAT\citep{Stanley2002}, and OVA-NEAT\citep{McDonnell}.\newline }
\label{tab:comp}
\begin{tabular}{p{0.09\textwidth}|p{0.14\textwidth}|p{0.12\textwidth}|p{0.12\textwidth}|p{0.135\textwidth}|p{0.12\textwidth}|p{0.12\textwidth}|p{0.11\textwidth}} \hline
Dataset & Method & RMSE\textsubscript{n} & RMSE\textsubscript{t} & Acc.\textsubscript{t} & \#H.L. & \#Neu. & \#Fea.\\ \hline 
\multirow{4}{*}{Flame} & GE & 0.28$\pm$0.15 & 0.31$\pm$0.15 & 0.884$\pm$0.11 & 1.00$\pm$0.00& 3.33$\pm$1.40& 1.97$\pm$0.18 \\ 
& SGE & 0.16$\pm$0.13 & 0.22$\pm$0.13 & 0.933$\pm$0.09 & 1.00$\pm$0.00& 4.47$\pm$1.83& 2.00$\pm$0.00\\ 
& DSGE & 0.07$\pm$0.05 & 0.16$\pm$0.08 & 0.974$\pm$0.04 & 2.81$\pm$0.60&12.5$\pm$6.64&2.00$\pm$0.00 \\ 
& MGE & 0.10$\pm$0.02 & 0.11$\pm$0.02 & \textbf{0.982$\pm$0.01} & 1.00$\pm$0.00& 7.13$\pm$1.63& 2.00$\pm$0.00 \\ \hline 
\multirow{4}{*}{WDBC} & GE & 0.24$\pm$0.09 & 0.27$\pm$0.09 & 0.903$\pm$0.12 & 1.00$\pm$0.00&3.17$\pm$1.53 & 8.40$\pm$3.81\\
& SGE & 0.19$\pm$0.03 & 0.23$\pm$0.04 & 0.932$\pm$0.02& 1.00$\pm$0.00& 3.73$\pm$1.53& 12.0$\pm$6.51 \\ 
& DSGE & 0.07$\pm$0.04 & 0.20$\pm$0.04 & \textbf{0.954$\pm$0.02} & 2.95$\pm$0.52& 12.8$\pm$6.70& 18.2$\pm$6.91\\ 
& MGE & 0.10$\pm$0.02 & 0.18$\pm$0.03 & \textbf{0.967$\pm$0.02} & 1.00$\pm$0.00& 9.00$\pm$1.63& 13.57$\pm$2.5 \\ \hline 
\multirow{4}{*}{Iono.} & GE & 0.33$\pm$0.10 & 0.38$\pm$0.07 & 0.764$\pm$0.18 & 1.00$\pm$0.00& 2.50$\pm$1.41& 7.33$\pm$5.33 \\ 
& SGE & 0.21$\pm$0.07 & 0.32$\pm$0.05 & 0.874$\pm$0.10 & 1.00$\pm$0.00& 3.53$\pm$1.36& 12.1$\pm$5.79\\
& DSGE & 0.14$\pm$0.05 & 0.29$\pm$0.05 & 0.898$\pm$0.03 & 2.50$\pm$0.51& 9.33$\pm$5.24& 15.8$\pm$5.46 \\ 
& MGE & 0.14$\pm$0.03 & 0.27$\pm$0.03 & \textbf{0.917$\pm$0.02} & 1.00$\pm$0.00& 9.53$\pm$1.87& 14.40$\pm$3.7 \\ \hline 
\multirow{4}{*}{Sonar} & GE & 0.38$\pm$0.08 & 0.45$\pm$0.06 & 0.681$\pm$0.12 & 1.00$\pm$0.00& 2.53$\pm$1.20& 9.40$\pm$5.73 \\ 
& SGE & 0.34$\pm$0.06 & 0.44$\pm$0.04 & 0.729$\pm$0.09 & 1.00$\pm$0.00 & 3.07$\pm$1.39 & 13.3$\pm$6.42 \\
& DSGE & 0.25$\pm$0.00 & 0.43$\pm$0.06 & \textbf{0.784$\pm$0.06} & 2.95$\pm$1.12& 10.9$\pm$4.12& 24.0$\pm$7.54 \\ 
& MGE & 0.28$\pm$0.01 & 0.39$\pm$0.02 & \textbf{0.793$\pm$0.03} & 1.00$\pm$0.00& 11.50$\pm$2.59& 17.07$\pm$3.5\\ \hline 
\multirow{4}{*}{Heart} & GE & 0.29$\pm$0.09 & 0.39$\pm$0.14 & 0.784$\pm$0.2 & 1.00$\pm$0.00 & 3.30$\pm$1.4 & 6.31$\pm$2.08\\ 
& SGE & 0.36$\pm$0.04 & 0.41$\pm$0.04 & 0.774$\pm$0.06 & 1.00$\pm$0.00& 5.73$\pm$2.40& 7.40$\pm$3.10 \\ 
& DSGE & 0.24$\pm$0.03 & 0.39$\pm$0.04 & 0.803$\pm$0.04 & 2.63$\pm$0.43& 7.67$\pm$2.30& 9.67$\pm$1.60\\ 
& MGE & 0.24$\pm$0.02 & 0.36$\pm$0.03 & \textbf{0.827$\pm$0.03} & 1.00$\pm$0.00& 7.40$\pm$2.00 & 6.93$\pm$1.23 \\ \hline 
\multirow{4}{*}{Credit} & GE & 0.33$\pm$0.01 & 0.44$\pm$0.10 & 0.721$\pm$0.21 & 1.00$\pm$0.00& 3.40$\pm$1.00 & 12.70$\pm$2.4 \\ 
& SGE & 0.47$\pm$0.04 & 0.47$\pm$0.04 & 0.643$\pm$0.11 & 1.00$\pm$0.00& 3.60$\pm$1.70& 5.03$\pm$3.12 \\ 
& DSGE & 0.42$\pm$0.11 & 0.44$\pm$0.03 & 0.664$\pm$0.05 & 2.83$\pm$1.02& 5.60$\pm$1.96&10.60$\pm$1.7 \\
& MGE & 0.30$\pm$0.02 & 0.41$\pm$0.02 & \textbf{0.754$\pm$0.02} & 1.00$\pm$0.00& 10.53$\pm$2.91& 15.10$\pm$2.7 \\ \hline 
\multirow{4}{*}{Wine} & GE & 0.26$\pm$0.06 & 0.34$\pm$0.16 & 0.889$\pm$0.27 & 1.00$\pm$0.00 & 2.81$\pm$1.00 & 7.67$\pm$1.82 \\ 
& SGE & 0.41$\pm$0.09 & 0.44$\pm$0.08 & 0.753$\pm$0.07 & 1.00$\pm$0.00& 5.06$\pm$2.49&7.13$\pm$2.67 \\ 
& DSGE & 0.17$\pm$0.06 & 0.34$\pm$0.09 & 0.864$\pm$0.07 & 2.30$\pm$0.86& 7.10$\pm$2.66&9.03$\pm$2.01 \\
& MGE & 0.15$\pm$0.03 & 0.31$\pm$0.02 & \textbf{0.934$\pm$0.04} & 1.00$\pm$0.00& 10.67$\pm$1.27& 8.06$\pm$1.26 \\ \hline 
\multirow{4}{*}{Digits} & GE & 0.24$\pm$0.03 & 0.31$\pm$0.04 & 0.798$\pm$0.02 & 1.00$\pm$0.00& 20.1$\pm$1.85 & 38.17$\pm$4.8 \\ 
& NEAT &  \multicolumn{1}{c|}{N/A} &  \multicolumn{1}{c|}{N/A} & 0.600$\pm$0.005 &  \multicolumn{1}{c|}{N/A} & \multicolumn{1}{c|}{51.00}&  \multicolumn{1}{c}{N/A}\\ 
& OVA-NEAT &  \multicolumn{1}{c|}{N/A} &  \multicolumn{1}{c|}{N/A} & \textbf{0.939$\pm$0.001} &  \multicolumn{1}{c|}{N/A} & \multicolumn{1}{c|}{399.00}&  \multicolumn{1}{c}{N/A} \\
& MGE & 0.18$\pm$0.03 & 0.27$\pm$0.01 & \textbf{0.945$\pm$0.02} & 1.00$\pm$0.00& 55.5$\pm$2.32& 59.80$\pm$3.5 \\ \hline
\multirow{4}{*}{Letter} & GE & 0.80$\pm$0.07 & 0.84$\pm$0.08 & 0.201$\pm$0.02 & 1.00$\pm$0.00& 17.5$\pm$2.01 & 14.60$\pm$1.3 \\ 
& NEAT &  \multicolumn{1}{c|}{N/A} &  \multicolumn{1}{c|}{N/A} & 0.253$\pm$0.001 &  \multicolumn{1}{c|}{N/A} & \multicolumn{1}{c|}{79.00}&  \multicolumn{1}{c}{N/A} \\ 
& OVA-NEAT & \multicolumn{1}{c|}{N/A}&  \multicolumn{1}{c|}{N/A} & 0.484$\pm$0.001 &  \multicolumn{1}{c|}{N/A} & \multicolumn{1}{c|}{560.00}&  \multicolumn{1}{c}{N/A}\\
& MGE & 0.55$\pm$0.03 & 0.61$\pm$0.01 & \textbf{0.547$\pm$0.02} & 1.00$\pm$0.00& 74.47$\pm$4.93& 15.03$\pm$1.0 \\ \hline 
\multirow{4}{*}{MNIST} & GE & 0.76$\pm$0.03 & 0.80$\pm$0.03 & 0.301$\pm$0.03 & 1.00$\pm$0.00 & 15.7$\pm$2.53 & 25.31$\pm$1.9 \\ 
& NEAT &  \multicolumn{1}{c|}{N/A} &  \multicolumn{1}{c|}{N/A} & 0.472$\pm$0.003 &  \multicolumn{1}{c|}{N/A} & \multicolumn{1}{c|}{94.00}&  \multicolumn{1}{c}{N/A} \\ 
& OVA-NEAT &  \multicolumn{1}{c|}{N/A} &  \multicolumn{1}{c|}{N/A} & 0.806$\pm$0.001 &  \multicolumn{1}{c|}{N/A} & \multicolumn{1}{c|}{337.00}&  \multicolumn{1}{c}{N/A} \\
& MGE & 0.30$\pm$0.02 & 0.35$\pm$0.02 & \textbf{0.893$\pm$0.02} & 1.00$\pm$0.00& 63.93$\pm$4.62& 177.1$\pm$11. \\ \hline 
\end{tabular}
\end{table*}

\noindent{\bf Training Performance}.\  MGE's RMSE\textsubscript{n} is more favorable than those of GE and SGE for all datasets. 
However, DSGE has a superior RMSE\textsubscript{n} in 3 cases. 
Statistical tests show that Acc.\textsubscript{t} of MGE is better than that of GE and SGE for all datasets with significant distance while in comparison with DSGE, the Acc.\textsubscript{t} of MGE is better in 5 cases with significant distance and is equal to DSGE's in 2 cases statistically. 
In addition, MGE's classification accuracy on test data is better than GE and NEAT for last three datasets (Digits, Letter, and MNIST) with significant distance.
MGE's test accuracy is equal statistically to that of OVA-NEAT for Digits dataset while it returns statistically better classification accuracy than OVA-NEAT's for Letter and MNIST datasets.

Similar to GE and SGE, MGE generates single-layered neural networks, while other methods (DSGE, NEAT, and OVA-NEAT) generate multi-layered networks. 
The single-layered topology directly affects  the MGE’s good generalization results demonstrated in Table 3. 
In addition, the single-layered neural networks are simpler to understand, easier to implement in hardware (in special applications) and faster to run in parallel environments. 
MGE's generated networks contain more neurons than those of GE, SGE and DSGE, because MGE reduces the mapping bias of GE towards simpler networks. 
In comparison with NEAT, MGE's output networks include slightly less hidden neurons for both of Letter and MNIST datasets, and  more hidden neurons for Digits dataset. 
As we expect, the networks generated by OVA-NEAT are much more complex than that of GE, NEAT and MGE.

\noindent{\bf Run Time Costs}.\ The number of fitness function calls during the evolution is a good measure of time complexity, due to the fact that it is the most time-consuming step in a generation. 
We can calculate the number of fitness function calls for each algorithm by multiplying the population size and the number of generations reported for the algorithm. 
Our algorithm terminates after 100000 evaluations for binary and ternary class datasets (the first 7 datasets) 
while GE and SGE use 50000 evaluations for the first 4 datasets and 100000 for the rest (50000 evaluations is hardly enough for the last three datasets). 
DSGE runs for 250000 evaluations on average. Thus, DSGE uses fairly more computational effort. 
Although GE and SGE take less computational costs in 4 cases (Flame, WDBC, Iono., and Sonar datasets), their classification accuracies are  weaker than MGE's. 
For fair comparisons, we re-run MGE on Flame, WDBC, Iono., and Sonar for 50000 evaluations (i.e., for 250 generations) and obtained respectively 0.978, 0.957, 0.902, and 0.771 classification accuracies that are again superior to that of GE and SGE with a significant distance. 
About the datasets with a large number of classes (i.e., Digits, Letter, and MNIST) NEAT uses respectively 449800, 240200, and 209200 evaluations and OVA-NEAT uses respectively 733400, 639400, and 575400 evaluations. By contrast,   MGE needs 600000 evaluations, which  is close to OVA-NEAT's evaluation count. 
To compare MGE with NEAT, we run MGE on Digits, Letter, and MNIST datasets for evaluations equal to that of NEAT, 
that is respectively for 449800\textbackslash200, 240200\textbackslash200, and 240200\textbackslash200 generations 
and obtained 0.931, 0.536, and 0.724 classification accuracies, which are  significantly better than NEAT's results (close to what we reported in Table \ref{tab:comp}).

\noindent{\bf Note}. Appendix A presents the evolutionary behavior of these algorithms in terms of the error curve of the fittest individual during evolutionary process.

\section{Comparisons with Other Classifiers}
\label{sec:exper3}

This section compares MGE's products with other classifiers namely k-Nearest Neighbor (k-NN), Classification and Regression Tree (CART) as a decision tree induction method, Support Vector Machines (SVM), Fully-connected Neural Network (FNN), and CNN in terms of classification accuracy and the classifier's complexity. 
We calculate the complexity by counting the number of basic floating-point operations that a generated classifier requires to predict an example's class.
Table \ref{tab:classifiers} compares the classifiers returned by MGE with those that \cite{Sun2021} generate by k-NN and  CART as well as with the classifiers that \cite{Zhang2020} obtain using SVM and Optimal Margin Distribution Machine (ODM).
\cite{Sun2021} demonstrate that their proposed feature selection applied before CART and k-NN improves the classification accuracy. We report both results with and without feature selection in Table \ref{tab:classifiers}.  In Table \ref{tab:classifiers}, the FS column shows if an algorithm uses feature selection or not. \cite{Sun2021} use ten-fold cross-validation to estimate the generalization ability of the methods. In a ten-fold cross-validation, the entire dataset is divided into 10 equal sized subsets, then in the run {\it i} of algorithm (1\ensuremath{\leq}i\ensuremath{\leq}10) the {\it i}-th subset is regarded as the testing set and the remaining 9 subsets of data as the training set. 
Table \ref{tab:classifiers} reports the average classification accuracy on testing data that CART and k-NN obtain from these 10 runs.
In Table \ref{tab:classifiers}, the MGE-cv record contains the results of MGE with ten-fold cross-validation setup.
\cite{Zhang2020} proposes ODM as a special type of SVM where almost the entire training set form support vectors. 
Table \ref{tab:classifiers} includes the results of SVM and ODM, both with (Radial Basis Function) RBF kernels on WDBC, Sonar, and Credit German datasets. 
In \cite{Zhang2020}'s experiment setup, 80\% of dataset patterns are randomly drawn as the training set and the remaining patterns form the testing set in each run of the algorithm. After 30 independent runs, the average classification accuracy on testing data for these methods (SVM and ODM) is reported. In Table \ref{tab:classifiers}, MGE-80-20 refers to the results of MGE using this 30x(80\%-20\%) experiment setup. 

\begin{table*}[h]
\center
\setlength\tabcolsep{1pt}
\caption{Comparisons between MGE and other classifiers (k-NN \citep{Sun2021}, CART \citep{Sun2021}, SVM \citep{Zhang2020}, and ODM \citep{Zhang2020}) in terms of classification accuracy and floating-point operations required to predict an example in the format of {\it ACC.\textsubscript{operations}}, where ACC denotes the accuracy and the subscript 'operations' represents the computational cost in terms of the number of floating-point operations. FS column indicates whether a classifier uses feature selection as the pre-processing step or not. The results of k-NN, CART, and MGE-cv are obtained from 10-fold cross-validation, while the results of SVM, ODM, and MGE-80-20 are the averages of 30 runs of these algorithms where the datasets are partitioned into 80\%-20\% as train-test sets in each run.\newline }
\label{tab:classifiers}
\begin{tabular}{p{0.14\textwidth}|P{0.04\textwidth}|p{0.11\textwidth}|p{0.11\textwidth}|p{0.11\textwidth}|p{0.11\textwidth}|p{0.11\textwidth}|p{0.11\textwidth}} \hline
Method & FS & WDBC & Iono. & Sonar & Heart & Credit & Wine\\ \hline 
\multirow{2}{*}{k-NN} & n & 0.912\textsubscript{15360} & 0.825\textsubscript{10710} & 0.865\textsubscript{11220} &0.752\textsubscript{3159} &  \multicolumn{1}{c|}{N/A}& 0.919\textsubscript{2080} \\
& y & 0.928\textsubscript{4153} & 0.903\textsubscript{4015} & 0.885\textsubscript{3500} &0.833\textsubscript{1543} &  \multicolumn{1}{c|}{N/A} & 0.972\textsubscript{1139} \\ \hline 
\multirow{2}{*}{CART} & n & 0.860\textsubscript{9} & 0.865\textsubscript{9} & 0.712\textsubscript{7}&0.785\textsubscript{8} &  \multicolumn{1}{c|}{N/A} & 0.893\textsubscript{7} \\
& y & 0.931\textsubscript{9} & 0.912\textsubscript{9} & 0.779\textsubscript{7}&0.782\textsubscript{8} & \multicolumn{1}{c|}{N/A} & 0.933\textsubscript{7}\\ \hline 
SVM  & n & 0.959 &  \multicolumn{1}{c|}{N/A} & 0.761&  \multicolumn{1}{c|}{N/A} & 0.737& \multicolumn{1}{c}{N/A} \\ \hline 
ODM & n & 0.969\textsubscript{15360} &  \multicolumn{1}{c|}{N/A} & 0.754\textsubscript{11220}&  \multicolumn{1}{c|}{N/A} & 0.744\textsubscript{2080}&  \multicolumn{1}{c}{N/A} \\ \hline
MGE-cv & n & 0.972\textsubscript{110} & 0.926\textsubscript{144}& 0.772\textsubscript{174} & 0.819\textsubscript{130} & 0.743\textsubscript{160} &0.978\textsubscript{146}\\ 
MGE-80-20 & n & 0.969\textsubscript{118} & 0.911\textsubscript{156}& 0.818\textsubscript{180} & 0.836\textsubscript{122} & 0.753\textsubscript{166} &0.935\textsubscript{140}\\ \hline
\end{tabular}
\end{table*} 

However, neither \citep{Sun2021} nor \citep{Zhang2020} report the floating-point operations required to predict an example.
Thus, we have calculated the criteria for them using other clues about their methods as follows:
First, we know that in a linear representation of training set as the most efficient implementation for our datasets, the nearest neighbor search requires $\mathcal{O}(n\times d)$ (floating-point) operations, where {\it n} and {\it d} are respectively the number of training examples and the number of input features. 
Thus, by assuming the number of floating-point operations to be $(n\times d)$, Table \ref{tab:classifiers} reports the operations count for 1-nearest neighbor search.
Second, \cite{WITTEN2011} estimate that decision tree's depth is $\mathcal{O}(log(n))$ based on the assumption that the induced tree is `bushy' and binary, where $n$ is the number of training examples. Accordingly, we assume that a prediction using CART's induced decision tree exactly takes $log(n)$ floating-point operations, and have obtained and reported the prediction cost of CART for datasets of Table \ref{tab:classifiers}.
Third, SVM takes $\mathcal{O}(n_{sv}\times d)$ (floating-point) operations for each classification, where {\it n\textsubscript{sv}} is the support vectors count \citep{Maji2013}. As the number of support vectors for SVM experiments is unreported in \citep{Zhang2020}, we left the computation cost of SVM blanked in the table. According to \citep{Zhang2020}, ODM uses almost the entire training set as support vectors that enables us to estimate the cost of ODM at least equal to $n_{sv}\times d$.
Fourth and finally, our experiments indicate that MGE's generated neural networks almost include 36, 44, 60, 40, 52, and 55 connections in average for WDBC, Ionosphere, Sonar, Heart, Credit, and Wine respectively.
When a neural network executes, each connection requires a couple of floating-point operations: one to calculate the product of input value and connection's weight and the other to sum up the products in the neuron, 
while each neuron's sigmoid activation function requires 4 floating-point operations based on the approximate method that \citep{Zhang1996} propose.
Thus, $2*a_1+4*a_2$ is the number of floating-point operations that a neural network 
with {\it a\textsubscript{1}} connections and {\it a\textsubscript{2}} neurons takes to execute. 
For example, the neural network of Figure \ref{fig:individual} containing 15 connections and 7 sigmoidal neurons takes $2*15 + 4*7 = 58$ floating-point operations to classify an input pattern.
Table \ref{tab:classifiers} includes the computation cost of MGE's generated ANNs based on this formula.

Table \ref{tab:classifiers} depicts that our classification accuracies are in the same order of other classifiers on the selected datasets, and MGE is the winner in most of the one-by-one comparisons. 
Since the competent methods only report their average values, we are unable to do statistical test. In terms of the number of floating-point operations to run a classifier, MGE's products are up to two orders of magnitude simpler than those of k-NN and ODM, while CART's induced trees are the fastest.

Finally, we compare the results of linear SVM, linear ODM, RBF-based SVM, Fully-Connected one layer NN containing 300 hidden neurons (FNN), and CNN (LeNet-4) with those of MGE on MNIST dataset.
Table \ref{tab:classifiers2} compares these methods in terms of the classification accuracy, the number of floating-point operations that each classification takes, and the dependence of the methods to the previous knowledge.
The classification accuracy of linear SVM and linear ODM are obtained from \citep{Zhang2020}, while the both of accuracy and complexity of RBF-based SVM, FNN, and CNN are obtained from \citep{Lecun1998}.
We are unable to estimate the complexity of linear SVM as the support vectors count of this method is unreported, while for ODM almost the entire training set are support vectors. As aforementioned above, we consider that ODM's classifier exactly takes $n_{sv}\times d$ floating-point operations for a prediction, where $n_{sv}=60,000$ and $d=784$ for MNIST dataset which means that ODM's classifier complexity would be close to 47,000,000.
The results demonstrate that the accuracy of MGE's generated ANNs is inferior to those of others. The neural networks generated by MGE are simpler than other generated classifiers, since ODM takes 47,000,000 floating-point operations per recognition, RBF-based SVM, FNN and LeNet-4 respectively take 650,000, 235800 and 260000 floating-point operations per recognition while MGE takes only 500 floating-point operations to classify a pattern. 
Furthermore, CNN is a human designed architecture style which is tuned for image classification tasks and both of CNN and FNN require significant try-and-error efforts to design a suitable architecture, while MGE automatically generates the both of architecture and weights.

\begin{table*}[h]
\center
\setlength\tabcolsep{1pt}
\caption{Comparisons among MGE, linear SVM \citep{Zhang2020}, linear ODM \citep{Zhang2020}, RBF-based SVM \citep{Lecun1998}, FNN \citep{Lecun1998}, and CNN \citep{Lecun1998} in terms of classification accuracy, floating-points operations that a classifier requires (denoted as Complexity), and the independence of the method to the human expertise.\newline}
\label{tab:classifiers2}
\begin{tabular}{p{0.20\textwidth}|P{0.12\textwidth}|P{0.16\textwidth}|P{0.16\textwidth}} \hline
Method  & Accuracy & Complexity & Is Automatic \\ \hline
Linear SVM & 0.925 &  \multicolumn{1}{c|}{N/A} &  y \\
Linear ODM& 0.952 & 47,000,000 &  y \\ 
RBF-based SVM  &  0.989 & 650,000 & y  \\
FNN&   0.953 & 235,800  & n \\  
CNN (LeNet-4)  & 0.989 &   260,000 & n  \\ 
MGE & 0.902 & 500 & y \\ \hline
\end{tabular}
\end{table*}

\FloatBarrier
\section{Conclusions}
\label{sec:conclusions}
In this paper, we proposed a Modular GE-based method (MGE) to automatically generate 
neural networks with modular topologies. 
The modular representation of MGE alleviates three major drawbacks of GE namely, weak locality, low scalability, and invalidity of individuals, which enables MGE to efficiently evolve  large ANNs. 
We hypothesized that if we restrict the solution space to modular neural networks, MGE generates more accurate ANNs. 
We defined five different forms of topological structures with/without modularity for ANNs and tested them empirically on four multi-class datasets (Wine, Digits, Letter, and MNIST). 
The results support a form of modularity with no inter-module connections (zero-coupling) where the hidden neurons forming a module are aligned in one layer.
Moreover, our experiments on binary class datasets  demonstrate that single-layer ANNs perform better than multi-layer ones.
Then, we compared MGE's outputs with five state-of-the-art NeuroEvolution algorithms (GE, SGE, DSGE, NEAT, and OVA-NEAT) using  ten classification benchmarks. 
The accuracy of generated ANNs on testing data is statistically better than the results of GE, SGE, DSGE, and NEAT in most of the cases (except for WDBC and Sonar that are equal statistically to DSGE's ones). 
OVA-NEAT, as an ensemble NeuroEvolution method, returns inferior accuracy compared with MGE for the Letter and MNIST datasets with a significant distance, while it returns accuracy statistically equivalent to MGE on Digits dataset.
Moreover, our generated networks are single-layered, which are preferable to the networks generated by DSGE, NEAT, and OVA-NEAT when deployed in resource-constrained devices. 
MGE's networks have fairly less neurons than that of OVA-NEAT in all comparisons.
In comparison with NEAT, MGE generates smaller networks for Letter and MNIST datasets and generates larger networks for Digits.
Then, we empirically compared the MGE's products with the outputs of some well-known machine learning methods.
On binary class datasets, the MGE's generated neural networks are comparable with the results of 1-nearest neighbor, CART, SVM, and ODM in terms of classification accuracy and complexity.
On MNIST dataset, we demonstrated that our networks are computationally simpler than SVM, ODM, FNN, and CNN up to several orders of magnitudes at the expense of a slight decrease in accuracy. 
Finally, our static representation analyses indicated that MGE (i) alleviates invalidity problem of
GE without any extra computation/memory overhead; (ii) has the strongest locality, and (iii) 
scales to more difficult and demanding problems compared to GE, SGE, and DSGE.
Moreover, MGE exhibits a stable locality unlike GE, SGE, and DSGE, whose locality decreases when the candidate solutions' complexity increases.


\textbf{Limitations:} First,   MGE encodes weights using GE representation that prevents us from performing tailored genetic operators for this type of parameters (like Gaussian mutation or median crossovers). 
Second, generalizing MGE to be used in other types of problems such as symbolic regression, path planning, and distributed program synthesis would be an interesting direction to pursue. 
Third, a subtle change in the mapping procedure enables MGE to generate recurrent neural networks; as an intuition, if we initially insert the index of all presenting hidden neurons in the genotype into the CFG, the $i$-th neuron is able to get connections count from the $j$-th neuron where $i\leq j$ and yet the $j$-th neuron can get connections from the $i$-th one. The result is an architecture with recurrent connections or loops. Fourth, we pose a question as: {\it How much the sparsity that the GE methods enforce on the output ANNs affect the performance of such algorithms in NeuroEvolution tasks?}

\textbf{Acknowledgement:} The authors would like to thank the anonymous reviewers for their insightful comments that helped us in improving this manuscript.

\small

\bibliographystyle{apalike}
\bibliography{sample-bibliography}

\begin{thebibliography}{}

\bibitem[Ahmadizar et~al., 2015]{ahmadizar2015}
Ahmadizar, F., Soltanian, K., AkhlaghianTab, F., and Tsoulos, I. (2015).
\newblock Artificial neural network development by means of a novel combination
  of grammatical evolution and genetic algorithm.
\newblock {\em Engineering Applications of Artificial Intelligence}, 39:1 --
  13.

\bibitem[Alford et~al., 2018]{Alford2018}
Alford, S., Robinett, R., Milechin, L., and Kepner, J. (2018).
\newblock Pruned and structurally sparse neural networks.
\newblock In {\em 2018 IEEE MIT Undergraduate Research Technology Conference
  (URTC)}, pages 1--4.

\bibitem[Amer and Maul, 2019]{Amer2019}
Amer, M. and Maul, T. (2019).
\newblock A review of modularization techniques in artificial neural networks.
\newblock {\em Artificial Intelligence Review}, 52(1):527--561.

\bibitem[Angeline and Pollack, 1993]{angeline1993}
Angeline, P.~J. and Pollack, J. (1993).
\newblock Evolutionary module acquisition.
\newblock In Fogel, D. and Atmar, W., editors, {\em Proceedings of the Second
  Annual Conference on Evolutionary Programming}, pages 154--163, La Jolla, CA,
  USA.

\bibitem[Angeline and Pollack, 1992]{angeline1992}
Angeline, P.~J. and Pollack, J.~B. (1992).
\newblock The evolutionary induction of subroutines.
\newblock In {\em Proceedings of the Fourteenth Annual Conference of the
  Cognitive Science Society}, pages 236--241, Bloomington, Indiana, USA.
  Lawrence Erlbaum.

\bibitem[Angeline et~al., 1994]{Angeline93}
Angeline, P.~J., Saunders, G.~M., and Pollack, J.~B. (1994).
\newblock An evolutionary algorithm that constructs recurrent neural networks.
\newblock {\em IEEE Transactions on Neural Networks}, 5(1):54--65.

\bibitem[Assun\c{c}\~{a}o et~al., 2017a]{Assuncao2017}
Assun\c{c}\~{a}o, F., Louren\c{c}o, N., Machado, P., and Ribeiro, B. (2017a).
\newblock Automatic generation of neural networks with structured grammatical
  evolution.
\newblock In {\em 2017 IEEE Congress on Evolutionary Computation (CEC)}, pages
  1557--1564.

\bibitem[Assun\c{c}\~{a}o et~al., 2017b]{Assunco-gecco}
Assun\c{c}\~{a}o, F., Louren\c{c}o, N., Machado, P., and Ribeiro, B. (2017b).
\newblock Towards the evolution of multi-layered neural networks: A dynamic
  structured grammatical evolution approach.
\newblock In {\em Proceedings of the Genetic and Evolutionary Computation
  Conference}, GECCO '17, pages 393--400, New York, NY, USA. Association for
  Computing Machinery.

\bibitem[Attwell and Laughlin, 2001]{Attwell2001}
Attwell, D. and Laughlin, S. (2001).
\newblock An energy budget for signaling in the grey matter of the brain.
\newblock {\em Journal of cerebral blood flow and metabolism : official journal
  of the International Society of Cerebral Blood Flow and Metabolism},
  21(10):1133—1145.

\bibitem[Auda and Kamel, 1999]{Gasser}
Auda, G. and Kamel, M. (1999).
\newblock Modular neural networks: a survey.
\newblock {\em International Journal of Neural Systems}, 9(2):129--151.

\bibitem[Azevedo et~al., 2009]{Frederico2009}
Azevedo, F. A.~C., Carvalho, L.~R., Grinberg, L., Farfel, J., Ferretti, R.~E.,
  Leite, R., Filho, W.~J., Lent, R., and Herculano-Houzel, S. (2009).
\newblock Equal numbers of neuronal and nonneuronal cells make the human brain
  an isometrically scaled-up primate brain.
\newblock {\em Journal of Comparative Neurology}, 513.

\bibitem[{Bartoli} et~al., 2020]{Bartoli}
{Bartoli}, A., {Castelli}, M., and {Medvet}, E. (2020).
\newblock Weighted hierarchical grammatical evolution.
\newblock {\em IEEE Transactions on Cybernetics}, 50(2):476--488.

\bibitem[Bengio et~al., 2006]{Bengio2006}
Bengio, Y., Lamblin, P., Popovici, D., and Larochelle, H. (2006).
\newblock Greedy layer-wise training of deep networks.
\newblock In {\em Proceedings of the 19th International Conference on Neural
  Information Processing Systems}, NIPS'06, page 153–160, Cambridge, MA, USA.
  MIT Press.

\bibitem[{Cantu-Paz} and {Kamath}, 2005]{Cantu}
{Cantu-Paz}, E. and {Kamath}, C. (2005).
\newblock An empirical comparison of combinations of evolutionary algorithms
  and neural networks for classification problems.
\newblock {\em IEEE Transactions on Systems, Man, and Cybernetics, Part B
  (Cybernetics)}, 35(5):915--927.

\bibitem[Chandra et~al., 2018]{Chandra2018}
Chandra, R., Gupta, A., Ong, Y.-S., and Goh, C.-K. (2018).
\newblock Evolutionary multi-task learning for modular knowledge representation
  in neural networks.
\newblock {\em Neural Processing Letters}, 47(3):993--1009.

\bibitem[Chen, 2015]{Chen2015}
Chen, K. (2015).
\newblock {\em Deep and Modular Neural Networks}, pages 473--494.
\newblock Springer Berlin Heidelberg, Berlin, Heidelberg.

\bibitem[Dheeru and Karra~Taniskidou, 2017]{UCI}
Dheeru, D. and Karra~Taniskidou, E. (2017).
\newblock {UCI} machine learning repository.

\bibitem[Eiben and Smith, 2015]{Eiben}
Eiben, A.~E. and Smith, J.~E. (2015).
\newblock {\em Introduction to Evolutionary Computing}.
\newblock Springer Publishing Company, Incorporated, 2nd edition.

\bibitem[Ellefsen et~al., 2020]{Ellefsen2019}
Ellefsen, K.~O., Huizinga, J., and Torresen, J. (2020).
\newblock Guiding neuroevolution with structural objectives.
\newblock {\em Evolutionary Computation}, 28(1):115--140.

\bibitem[Ellefsen et~al., 2015]{Ellefsen2015}
Ellefsen, K.~O., Mouret, J.-B., and Clune, J. (2015).
\newblock Neural modularity helps organisms evolve to learn new skills without
  forgetting old skills.
\newblock {\em PLoS computational biology}, 11(4):e1004128.

\bibitem[Fu and Medico, 2007]{Flame}
Fu, L. and Medico, E. (2007).
\newblock Flame, a novel fuzzy clustering method for the analysis of dna
  microarray data.
\newblock {\em BMC Bioinformatics}, 8(1):3.

\bibitem[{Galvan-Lopez} and {O'Neill}, 2009]{Galvan-Lopez}
{Galvan-Lopez}, E. and {O'Neill}, M. (2009).
\newblock On the effects of locality in a permutation problem: The sudoku
  puzzle.
\newblock In {\em 2009 IEEE Symposium on Computational Intelligence and Games},
  pages 80--87.

\bibitem[Glorot et~al., 2011]{Glorot2011}
Glorot, X., Bordes, A., and Bengio, Y. (2011).
\newblock Deep sparse rectifier neural networks.
\newblock In Gordon, G., Dunson, D., and Dudík, M., editors, {\em Proceedings
  of the Fourteenth International Conference on Artificial Intelligence and
  Statistics}, volume~15 of {\em Proceedings of Machine Learning Research},
  pages 315--323, Fort Lauderdale, FL, USA. PMLR.

\bibitem[Gottlieb and Eckert, 2000]{Gottlieb}
Gottlieb, J. and Eckert, C. (2000).
\newblock A comparison of two representations for the fixed charge
  transportation problem.
\newblock In Schoenauer, M., Deb, K., Rudolph, G., Yao, X., Lutton, E., Merelo,
  J.~J., and Schwefel, H.-P., editors, {\em Parallel Problem Solving from
  Nature PPSN VI}, pages 345--354, Berlin, Heidelberg. Springer Berlin
  Heidelberg.

\bibitem[Grisci et~al., 2019]{Grisci}
Grisci, B.~I., Feltes, B.~C., and Dorn, M. (2019).
\newblock Neuroevolution as a tool for microarray gene expression pattern
  identification in cancer research.
\newblock {\em Journal of Biomedical Informatics}, 89:122 -- 133.

\bibitem[Gruau et~al., 1996]{Gruau}
Gruau, F., Whitley, D., and Pyeatt, L. (1996).
\newblock A comparison between cellular encoding and direct encoding for
  genetic neural networks.
\newblock In {\em Proceedings of the 1st Annual Conference on Genetic
  Programming}, pages 81--89, Cambridge, MA, USA. MIT Press.

\bibitem[Hagg et~al., 2017]{Alexander}
Hagg, A., Mensing, M., and Asteroth, A. (2017).
\newblock Evolving parsimonious networks by mixing activation functions.
\newblock In {\em Proceedings of the Genetic and Evolutionary Computation
  Conference}, GECCO '17, pages 425--432, New York, NY, USA. Association for
  Computing Machinery.

\bibitem[{Harper} and {Blair}, 2006]{Harper}
{Harper}, R. and {Blair}, A. (2006).
\newblock Dynamically defined functions in grammatical evolution.
\newblock In {\em 2006 IEEE International Conference on Evolutionary
  Computation}, pages 2638--2645.

\bibitem[Hemberg~E., 2009]{Hemberg2009}
Hemberg~E., O'Neill~M., B.~A. (2009).
\newblock An investigation into automatically defined function representations
  in grammatical evolution.
\newblock In {\em Mendel 2009 15th International Conference on Soft Computing}.

\bibitem[Hinton et~al., 2006]{Hinton2006}
Hinton, G.~E., Osindero, S., and Teh, Y.-W. (2006).
\newblock A fast learning algorithm for deep belief nets.
\newblock {\em Neural Comput.}, 18(7):1527--1554.

\bibitem[{Khare} et~al., 2005]{Khare2005}
{Khare}, V.~R., {Xin Yao}, {Sendhoff}, B., {Yaochu Jin}, and {Wersing}, H.
  (2005).
\newblock Co-evolutionary modular neural networks for automatic problem
  decomposition.
\newblock In {\em IEEE Congress on Evolutionary Computation}, pages 2691--2698.

\bibitem[{Kim} and {Cho}, 2019]{T-Kim}
{Kim}, T. and {Cho}, S. (2019).
\newblock Particle swarm optimization-based cnn-lstm networks for forecasting
  energy consumption.
\newblock In {\em IEEE Congress on Evolutionary Computation (CEC)}, pages
  1510--1516.

\bibitem[Kitano, 1990]{kitano90}
Kitano, H. (1990).
\newblock Designing neural networks using genetic algorithms with graph
  generation system.
\newblock {\em Complex Systems}, 4:461--476.

\bibitem[Koza, 1994]{Koza1994}
Koza, J.~R. (1994).
\newblock {\em Genetic Programming II: Automatic Discovery of Reusable
  Programs}.
\newblock MIT Press, Cambridge, MA, USA.

\bibitem[Lecun et~al., 1998]{Lecun1998}
Lecun, Y., Bottou, L., Bengio, Y., and Haffner, P. (1998).
\newblock Gradient-based learning applied to document recognition.
\newblock {\em Proceedings of the IEEE}, 86(11):2278--2324.

\bibitem[Lennie, 2003]{Lennie2003}
Lennie, P. (2003).
\newblock The cost of cortical computation.
\newblock {\em Current Biology}, 13(6):493--497.

\bibitem[Li et~al., 2014]{Xiao}
Li, X.-L., Jia, C., Liu, D.-X., and Ding, D.-W. (2014).
\newblock Nonlinear adaptive control using multiple models and dynamic neural
  networks.
\newblock {\em Neurocomputing}, 136:190 -- 200.

\bibitem[Louren\c{c}o et~al., 2016]{Lourenco}
Louren\c{c}o, N., Pereira, F.~B., and Costa, E. (2016).
\newblock Unveiling the properties of structured grammatical evolution.
\newblock {\em Genetic Programming and Evolvable Machines}, 17(3):251--289.

\bibitem[Lu et~al., 2019]{Zhichao}
Lu, Z., Whalen, I., Boddeti, V., Dhebar, Y., Deb, K., Goodman, E., and Banzhaf,
  W. (2019).
\newblock Nsga-net: Neural architecture search using multi-objective genetic
  algorithm.
\newblock In {\em Proceedings of the Genetic and Evolutionary Computation
  Conference}, GECCO '19, pages 419--427, New York, NY, USA. Association for
  Computing Machinery.

\bibitem[Maji et~al., 2013]{Maji2013}
Maji, S., Berg, A.~C., and Malik, J. (2013).
\newblock Efficient classification for additive kernel svms.
\newblock {\em IEEE Transactions on Pattern Analysis and Machine Intelligence},
  35(1):66--77.

\bibitem[McDonnell et~al., 2018]{McDonnell}
McDonnell, T., Andoni, S., Bonab, E., Cheng, S., Choi, J.-H., Goode, J., Moore,
  K., Sellers, G., and Schrum, J. (2018).
\newblock Divide and conquer: Neuroevolution for multiclass classification.
\newblock In {\em Proceedings of the Genetic and Evolutionary Computation
  Conference}, GECCO '18, pages 474--481, New York, NY, USA. Association for
  Computing Machinery.

\bibitem[Medvet, 2017]{Medvet2017}
Medvet, E. (2017).
\newblock A comparative analysis of dynamic locality and redundancy in
  grammatical evolution.
\newblock In McDermott, J., Castelli, M., Sekanina, L., Haasdijk, E., and
  Garc{\'i}a-S{\'a}nchez, P., editors, {\em Genetic Programming}, pages
  326--342, Cham. Springer International Publishing.

\bibitem[Medvet et~al., 2019]{Medvet2019}
Medvet, E., Bartoli, A., De~Lorenzo, A., and Tarlao, F. (2019).
\newblock Designing automatically a representation for grammatical evolution.
\newblock {\em Genetic Programming and Evolvable Machines}, 20(1):37--65.

\bibitem[Medvet et~al., 2017]{Medvet17}
Medvet, E., Daolio, F., and Tagliapietra, D. (2017).
\newblock Evolvability in grammatical evolution.
\newblock In {\em Proceedings of the Genetic and Evolutionary Computation
  Conference}, GECCO '17, pages 977--984, New York, NY, USA. ACM.

\bibitem[Miikkulainen et~al., 2019]{Miikkulainen17}
Miikkulainen, R., Liang, J., Meyerson, E., Rawal, A., Fink, D., Francon, O.,
  Raju, B., Shahrzad, H., Navruzyan, A., Duffy, N., and Hodjat, B. (2019).
\newblock Chapter 15 - evolving deep neural networks.
\newblock In Kozma, R., Alippi, C., Choe, Y., and Morabito, F.~C., editors,
  {\em Artificial Intelligence in the Age of Neural Networks and Brain
  Computing}, pages 293 -- 312. Academic Press.

\bibitem[Miller, 2011]{Miller2011}
Miller, J.~F. (2011).
\newblock {\em Cartesian Genetic Programming}, pages 17--34.
\newblock Springer Berlin Heidelberg, Berlin, Heidelberg.

\bibitem[Miller, 2020]{Miller2019}
Miller, J.~F. (2020).
\newblock Cartesian genetic programming: its status and future.
\newblock {\em Genetic Programming and Evolvable Machines}, 21:129--168.

\bibitem[Nguyen, 2011]{Nguyen}
Nguyen, Q.~U. (2011).
\newblock {\em Examining Semantic Diversity and Semantic Locality of Operators
  in Genetic Programming}.
\newblock PhD thesis, University College Dublin, Ireland.

\bibitem[O'Neill et~al., 2008]{GEVA}
O'Neill, M., Hemberg, E., Gilligan, C., Bartley, E., McDermott, J., and
  Brabazon, A. (2008).
\newblock Geva: Grammatical evolution in java.
\newblock {\em SIGEVOlution}, 3(2).

\bibitem[O'Neill and Ryan, 1999]{Oniel99}
O'Neill, M. and Ryan, C. (1999).
\newblock Genetic code degeneracy: Implications for grammatical evolution and
  beyond.
\newblock In Floreano, D., Nicoud, J.-D., and Mondada, F., editors, {\em
  Advances in Artificial Life}, pages 149--153, Berlin, Heidelberg. Springer
  Berlin Heidelberg.

\bibitem[O'Neill and Ryan, 2000]{Oneill2000}
O'Neill, M. and Ryan, C. (2000).
\newblock Grammar based function definition in grammatical evolution.
\newblock In {\em Proceedings of the 2Nd Annual Conference on Genetic and
  Evolutionary Computation}, GECCO'00, pages 485--490, San Francisco, CA, USA.
  Morgan Kaufmann Publishers Inc.

\bibitem[Pawlik and Augsten, 2015]{Pawlik}
Pawlik, M. and Augsten, N. (2015).
\newblock Efficient computation of the tree edit distance.
\newblock {\em ACM Trans. Database Syst.}, 40(1):3:1--3:40.

\bibitem[Rothlauf and Goldberg, 2003]{Rothlauf}
Rothlauf, F. and Goldberg, D.~E. (2003).
\newblock Redundant representations in evolutionary computation.
\newblock {\em Evol. Comput.}, 11(4):381--415.

\bibitem[Rumelhart et~al., 1986]{Rumelhart}
Rumelhart, D.~E., Hinton, G.~E., and Williams, R.~J. (1986).
\newblock Learning representations by back-propagating errors.
\newblock {\em Nature}, 323:533--536.

\bibitem[Ryan et~al., 1998]{Ryan1998}
Ryan, C., Collins, J., and Neill, M.~O. (1998).
\newblock Grammatical evolution: Evolving programs for an arbitrary language.
\newblock In Banzhaf, W., Poli, R., Schoenauer, M., and Fogarty, T.~C.,
  editors, {\em Genetic Programming}, pages 83--96, Berlin, Heidelberg.
  Springer Berlin Heidelberg.

\bibitem[{Schrum} and {Miikkulainen}, 2016]{Schrum}
{Schrum}, J. and {Miikkulainen}, R. (2016).
\newblock Discovering multimodal behavior in ms.pac-man through evolution of
  modular neural networks.
\newblock {\em IEEE Transactions on Computational Intelligence and AI in
  Games}, 8(1):67--81.

\bibitem[Stanley and Miikkulainen, 2002]{Stanley2002}
Stanley, K.~O. and Miikkulainen, R. (2002).
\newblock Evolving neural networks through augmenting topologies.
\newblock {\em Evol. Comput.}, 10(2):99--127.

\bibitem[Sun et~al., 2021]{Sun2021}
Sun, L., Wang, L., Ding, W., Qian, Y., and Xu, J. (2021).
\newblock Feature selection using fuzzy neighborhood entropy-based uncertainty
  measures for fuzzy neighborhood multigranulation rough sets.
\newblock {\em IEEE Transactions on Fuzzy Systems}, 29(1):19--33.

\bibitem[Swafford et~al., 2011a]{Swafford2011}
Swafford, J.~M., Hemberg, E., O'Neill, M., Nicolau, M., and Brabazon, A.
  (2011a).
\newblock A non-destructive grammar modification approach to modularity in
  grammatical evolution.
\newblock In {\em Proceedings of the 13th Annual Conference on Genetic and
  Evolutionary Computation}, GECCO '11, pages 1411--1418, New York, NY, USA.
  ACM.

\bibitem[Swafford et~al., 2011b]{Swafford2011a}
Swafford, J.~M., O'Neill, M., Nicolau, M., and Brabazon, A. (2011b).
\newblock Exploring grammatical modification with modules in grammatical
  evolution.
\newblock In {\em Proceedings of the 14th European Conference on Genetic
  Programming}, EuroGP'11, pages 310--321, Berlin, Heidelberg. Springer-Verlag.

\bibitem[Thorhauer, 2016]{Thorhauer}
Thorhauer, A. (2016).
\newblock On the non-uniform redundancy in grammatical evolution.
\newblock In Handl, J., Hart, E., Lewis, P.~R., L{\'o}pez-Ib{\'a}{\~{n}}ez, M.,
  Ochoa, G., and Paechter, B., editors, {\em Parallel Problem Solving from
  Nature -- PPSN XIV}, pages 292--302, Cham. Springer International Publishing.

\bibitem[Tsoulos et~al., 2008]{tsoulos2008}
Tsoulos, I., Gavrilis, D., and Glavas, E. (2008).
\newblock Neural network construction and training using grammatical evolution.
\newblock {\em Neurocomputing}, 72(1):269--277.

\bibitem[{Watts} et~al., 2019]{Watts}
{Watts}, T., {Xue}, B., and {Zhang}, M. (2019).
\newblock Blocky net: A new neuroevolution method.
\newblock In {\em 2019 IEEE Congress on Evolutionary Computation (CEC)}, pages
  586--593.

\bibitem[Whiteson et~al., 2005]{Whiteson}
Whiteson, S., Stone, P., Stanley, K.~O., Miikkulainen, R., and Kohl, N. (2005).
\newblock Automatic feature selection in neuroevolution.
\newblock In {\em Proceedings of the 7th Annual Conference on Genetic and
  Evolutionary Computation}, GECCO '05, pages 1225--1232, New York, NY, USA.
  Association for Computing Machinery.

\bibitem[Whitley, 1989]{whitley89}
Whitley, D. (1989).
\newblock The genitor algorithm and selection pressure: Why rank-based
  allocation of reproductive trials is best.
\newblock In {\em Proceedings of the Third International Conference on Genetic
  Algorithms}, pages 116--121, San Francisco, CA, USA. Morgan Kaufmann
  Publishers Inc.

\bibitem[Witten et~al., 2011]{WITTEN2011}
Witten, I.~H., Frank, E., and Hall, M.~A. (2011).
\newblock Chapter 6 - implementations: Real machine learning schemes.
\newblock In Witten, I.~H., Frank, E., and Hall, M.~A., editors, {\em Data
  Mining: Practical Machine Learning Tools and Techniques (Third Edition)}, The
  Morgan Kaufmann Series in Data Management Systems, pages 191--304. Morgan
  Kaufmann, Boston, third edition edition.

\bibitem[Yaman et~al., 2018]{Anil2018}
Yaman, A., Mocanu, D.~C., Iacca, G., Fletcher, G., and Pechenizkiy, M. (2018).
\newblock Limited evaluation cooperative co-evolutionary differential evolution
  for large-scale neuroevolution.
\newblock In {\em Proceedings of the Genetic and Evolutionary Computation
  Conference}, GECCO '18, pages 569--576, New York, NY, USA. Association for
  Computing Machinery.

\bibitem[Zhang et~al., 1996]{Zhang1996}
Zhang, M., Vassiliadis, S., and Delgado-Frias, J. (1996).
\newblock Sigmoid generators for neural computing using piecewise
  approximations.
\newblock {\em IEEE Transactions on Computers}, 45(9):1045--1049.

\bibitem[Zhang and Zhou, 2020]{Zhang2020}
Zhang, T. and Zhou, Z.-H. (2020).
\newblock Optimal margin distribution machine.
\newblock {\em IEEE Transactions on Knowledge and Data Engineering},
  32(6):1143--1156.

\bibitem[{Zoph} et~al., 2018]{Zoph2018}
{Zoph}, B., {Vasudevan}, V., {Shlens}, J., and {Le}, Q.~V. (2018).
\newblock Learning transferable architectures for scalable image recognition.
\newblock In {\em 2018 IEEE/CVF Conference on Computer Vision and Pattern
  Recognition}, pages 8697--8710.

\end{thebibliography}

\normalsize
\newpage

\appendix
\section*{APPENDIX}

\section{Error Curve of Fittest Individual}
\label{sec:appendix1}

\begin{figure*}[h]
		\includegraphics[width=0.99\textwidth]{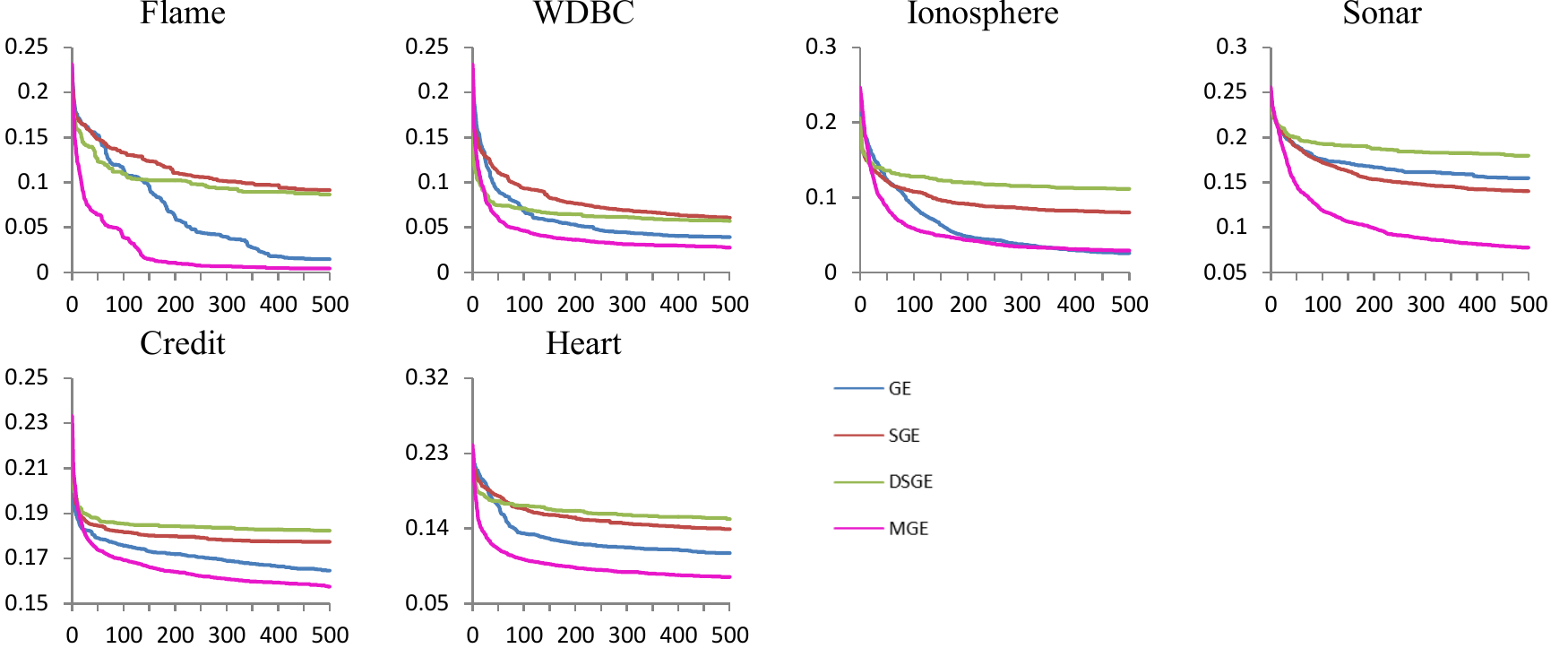}
		\caption{Error minimization curve of the best individual through evolution of GE, SGE, DSGE and MGE during 500 generations where the population size is 1000.}
		\label{fig:evolutioncurve}
\end{figure*}

The curve of error for the fittest individual during evolution may provide important information about the algorithm's efficiency by disclosing the speed of convergence and the steepness of the curve towards the end of run \citep{Eiben}. It is favorable for an algorithm to converge faster, but it is undesirable for its curve to flatten out early. 
Figure \ref{fig:evolutioncurve} illustrates the results for MGE, GE, SGE, and DSGE on binary-class datasets.
The size of the population in all of experiments is 1000 and we use Mean Square Error (MSE) criterion as the fitness function.
Observe that, MGE converges faster.
The notable outcome is that MGE's results are considerably different from that of SGE and DSGE on almost all of the datasets. 
Moreover, the MGE significantly differs from GE on Sonar, Credit, and Heart datasets, which are the hardest problems among these datasets. 
We conjecture that the decline of MGE's would be more rapid for harder problems, while the curves are steeper than or equal to other methods for almost all datasets.
Note that, MGE uses MSE as the fitness function (instead of cross-entropy) to be consistent with other GEs and also we have ignored the comparisons on multi-class datasets where the error of our modular neural networks would be clearly better than other GEs. 

\end{document}